\documentclass[journal]{IEEEtran}
\usepackage[utf8]{inputenc} 
\usepackage[T1]{fontenc}    
\usepackage{lineno, hyperref}       
\usepackage{url}            
\usepackage{booktabs}       
\usepackage{amsfonts}       
\usepackage{nicefrac}       
\usepackage{microtype}      
\usepackage{amsmath, amssymb, amsthm, amsfonts}
\usepackage{multirow}
\usepackage{threeparttable}
\usepackage{graphicx}
\usepackage{caption}
\usepackage{subfigure}
\usepackage{wrapfig}
\usepackage{color}
\usepackage{diagbox}
\usepackage{latexsym}
\usepackage{makecell}
\usepackage{threeparttable}

\begin{document}
%
\title{Inner-Imaging Networks: Put Lenses into Convolutional Structure}
%
%
%

\author{Yang Hu,~\IEEEmembership{Student Member,~IEEE,}
		Guihua Wen,
        Mingnan Luo,
		Dan Dai,
        Wenming Cao,
		Zhiwen Yu,~\IEEEmembership{Senior Member,~IEEE,}
        and Wendy Hall
\thanks{Yang Hu, Guihua Wen, MingnanLuo, Dan Dai, Zhiwen Yu are with the School of Computer Science and Engineering, South China University of Technology, China. Yang Hu is also with University of Southampton, UK. Email: superhy199148@hotmail.com, crghwen@scut.edu.cn, Phone no: +86-18998384808.}
\thanks{Wenming Cao is with the Department of Computer Science, City University of Hong Kong, Hong Kong. Email:wenmincao2-c@my.cityu.edu.hk.}
\thanks{Wendy Hall is with the Web Science Institute, University of Southampton, UK. Email:wh@ecs.soton.ac.uk, Phone no: +44(0)2380592388.}
\thanks{Guihua Wen is the corresponding author.}
}

%
%

\markboth{Submit to IEEE transactions}%
{Shell \MakeLowercase{\textit{et al.}}: Bare Demo of IEEEtran.cls for IEEE Journals}
%



\maketitle

\begin{abstract}
Despite the tremendous success in computer vision, deep convolutional networks suffer from serious computation costs and redundancies. Although previous works address that by enhancing the diversities of filters, they have not considered the complementarity and the completeness of the internal convolutional structure. To respond to this problem, we propose a novel Inner-Imaging architecture, which allows relationships between channels to meet the above requirement. Specifically, we organize the channel signal points in groups using convolutional kernels to model both the intra-group and inter-group relationships simultaneously. A convolutional filter is a powerful tool for modeling spatial relations and organizing grouped signals, so the proposed methods map the channel signals onto a pseudo-image, like putting a lens into the internal convolution structure. Consequently, not only the diversity of channels is increased, but also the complementarity and completeness can be explicitly enhanced. The proposed architecture is lightweight and easy to be implemented. It provides an efficient self-organization strategy for convolutional networks to improve their performance. Extensive experiments are conducted on multiple benchmark datasets, including CIFAR, SVHN, and ImageNet. Experimental results verify the effectiveness of the Inner-Imaging mechanism with the most popular convolutional networks as the backbones.
\end{abstract}


%
\IEEEpeerreviewmaketitle

\section{Introduction}
\label{sec1}
%
%
%
%
\IEEEPARstart{D}{eep} convolutional neural networks (CNNs) have exhibited significant effectiveness in modeling image data~\cite{szegedy2015going,wei2017cross,wu2017fuiqa,wu2018face,li2018tooth,hu2019automatic}; their structures have also been explored continuously~\cite{gu2018recent,he2016deep,zhang2018feature,wang2017building,wang2019enhancing}. Meanwhile, CNNs show the bulky size and severe redundancy~\cite{zhang2017polynet,larsson2017fractalnet:}. Besides pruning a complete structure~\cite{yu2018nisp:,zhu2017feature,zeng2018accelerating}, lots of methods aim to improve the efficiency of CNNs~\cite{sandler2018mobilenetv2,ma2018shufflenet}. Generally, the efficiency heavily depends on the interior components of CNNs, which should meet the following requirements: diversity, complementarity, and completeness. As the basic elements of CNNs, convolutional filters are often modeled to implement channel-wise attention~\cite{hu2018squeeze-and-excitation}, which only focuses on improving the diversity of feature maps, and lacks explicit modeling of complementarity and completeness of convolution channels.

Some methods have designed to model the grouping relationship between convolution channels~\cite{chollet2017xception,zhang2018shufflenet,huang2018condensenet}. Where, Xception~\cite{chollet2017xception} encodes the channels modeled by grouping, enhancing the interaction of features between groups, Shufflenet~\cite{zhang2018shufflenet} explicitly proposed the interaction and fusion of channels between different groups, and ConDenseNet~\cite{huang2018condensenet} further verified the excellent performance of convolutional channel-group modeling on the structure of DenseNet~\cite{huang2017densely}. Their outstanding performances indicate that implicit group relations exist between convolutional feature maps. However, the previous works failed to feedback on the modeling of the convolution channel-group relations to the optimization process of the feature maps. On the other hand, the grouping relations are ignored in the ordinary channel-wise attention methods. Since their plain Fully-connected (FC) encoder cannot represent the grouping and interaction of channel relationships. In other words, these methods have not explicitly modeled the coordination and complementarity between channels.

\begin{figure*}[!t]
\centering
\includegraphics[scale=0.78]{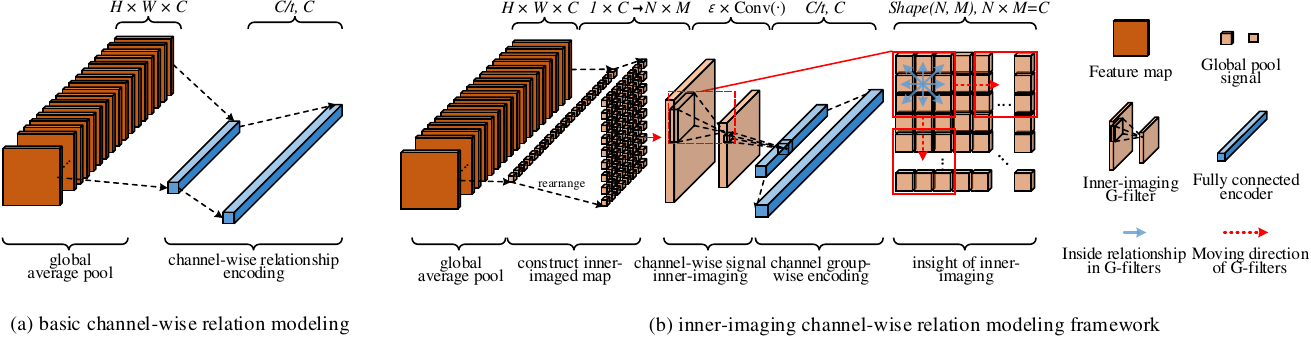}
\caption{Architectural comparison of channel relationship modeling. (a): The typical channel relation encoder~\cite{hu2018squeeze-and-excitation} with considering of identity mapping; (b): The Inner-Imaging channel-wise relationship modeling architecture, which organizes and models the channel relationships inside each group and between groups.}
\label{fig1}
\end{figure*}

To overcome the above shortcomings, this paper proposes a novel "Inner-Imaging" (InI) mechanism, as shown in Fig.\ref{fig1}(b), which is a new way to model the channel relationships. Compare with the channel-relationship modeling in~\cite{hu2018squeeze-and-excitation}(as shown in Fig.~\ref{fig1}(a)), our method first rearranges the feature signals into a pseudo-image $\hat{\mathrm{v}} \in \mathbb{R}^{N \times M}$, then it creates grouping filters (G-filters) $\mathrm{w}^{(a \times b)}$ to scan on it. In this process, channel signals ${u}_{ij} \in \hat{\mathrm{v}}$ within the same receptive field can build relations from multiple directions (top, down, left, right, top left, top right, bottom left, bottom right, and so on), and are assigned into one group. Subsequently, InI adopts FC layers to model group-wise relationships. That is, the G-filters are responsible for modeling relations between channels in the same group; the followed FC layers are used to model the relationships between groups. With this strategy, the complementarities of both convolutional channels and channel groups are enhanced. On the other hand, the size of channel groups can be flexibly controlled by adjusting the shape of G-filters (such as G-filters $\mathrm{w}^{(1 \times 1)}$ with the shape $1 \times 1$ representing the groups with the single-channel). In this way, the completeness of the representation of channels can be improved by integrating multi-scale G-filters. The InI-model provides a more complete and precise convolution channel relationship modeling method and provides the channels more rational re-scaling weights.

The exploration of CNN architecture and modeling of internal network representation is a meaningful and challenging task~\cite{liao2017a,zoph2018learning,pham2018efficient}. The design of "Inner-Imaging" brings a new idea of convolution internal structure modeling. It also provides us a carrier to explore grouping modes of convolution channels. The InI architecture can be applied to all kinds of CNNs to improve the efficiency of CNNs; it is lightweight and easy to implement and understand.

Our contributions can be summarized as follows:

\begin{itemize}
\item A novel "Inner-Imaging" (InI) mechanism is proposed, which first uses G-filters to organize channel signals, and simultaneously models the intra-group channel relationships and the inter-group channel relationships. Some theoretical deductions of InI are also provided.
\item The diversity of G-filters with different utility is explored. Moreover, it is also proposed that multi-shape G-filters can be integrated to realize the fusion of multi-size channel-group modeling.
\item The InI mechanism is employed in some popular CNN structures. For residual networks (ResNets), it builds inner-imaged maps with both residual and identity mappings, enabling identification flows to participate in the attentional process of residual flows.
\item With the InI mechanism, the effect of diverse channel grouping types is analyzed with the ablation studies for each mode of the InI-model. Besides, the ability of the InI module to collaborate with the spatial attention mechanism~\cite{chen2017sca-cnn:,woo2018cbam:} is verified.
\end{itemize}

The remainder of this paper is organized as follows. Section~\ref{sec2} discusses related works. Section~\ref{sec3} introduces the overall framework of the InI mechanism and its enhanced edition for residual networks. Section~\ref{sec4} presents our theoretical explanations for designing of the InI mechanism. Section~\ref{sec5} describes the experimental results and analysis. We conclude in Section~\ref{sec7}.

\section{Related Works}
\label{sec2}

\subsection{Efficient convolution structures}
\label{sec2.1}
Both huge volume and calculation of CNNs~\cite{cheng2017quantized,yu2018deep,tung2018deep,zagoruyko2016wide} are considered due to its serve redundancy~\cite{huang2016deep,veit2016residual}, which easily leads to inefficient modeling and over-fitting. Some methods impose regularization constraints to the network features~\cite{lu2018aar-cnns:,lin2019towards}, or random occlusion or perturbation of intermediate features~\cite{wang2016beyond,ghiasi2018dropblock}, some methods attempt to prune the block or channel for an over-complete convolutional architecture~\cite{wu2018blockdrop:,liu2018computation-performance,luo2018thinet:}, or use an early exit mechanism~\cite{figurnov2017spatially}. They apply destructive simplification to some complete models, rather than increase the modeling efficiency of finite-scale models. These methods need to build an initial large-scale network and consume computation to optimize pruning operations.

In a parallel line, efficient use of existing components and features is a two-pronged approach~\cite{zhao2018deep,huo2018training}. Some methods densely model the feature maps~\cite{huang2017densely,huang2018condensenet,yang2018convolutional}. To make CNN channels organized well, the modeling of channel relationships has attracted research attention~\cite{adebayo2018sanity}. Some approaches attempt to enhance the association of channels~\cite{gao2018channelnets:} and achieve high-efficiency performance by modeling them in the grouping~\cite{ma2018shufflenet,huang2018condensenet}. The studies mentioned above only refine the features of the middle layers repetitively or train the convolution channels in fixed groups, and they do not use the channel relationships to re-scale the feature maps. Compared with them, the InI mechanism can model the channel grouping relationships with various sizes and use the attention module to re-weight them.

\subsection{Attention and gating mechanisms in CNNs}
\label{sec2.2}
Diversified representation capability is a vital target pursued by machine learning models~\cite{cao2015diversity,luo2018consistent,yang2019split,zhang2018generalized}. Attention is widely applied to improve the diversity representation of CNNs~\cite{nguyen2018attentive}. It is typically used to model the spatial attentional area~\cite{li2018harmonious,li2018eac-net:,li2018deep,zhang2018medical} and content meaning~\cite{hou2018content-attention}, including multi-scale~\cite{chen2016attention,newell2016stacked} and multi-shape~\cite{jaderberg2015spatial,tang2017pixel} features. As a tool for biasing the allocation of resources~\cite{hu2018squeeze-and-excitation}, attention is also used to regulate the internal
CNN features~\cite{perez2018film:,stollenga2014deep}. Unlike channel switching, combination~\cite{zhang2018shufflenet:,wang2018revisiting}, channel-wise attention provides an end-to-end training solution for re-weighting the intermediate channel features. It can be also combined with spatial attention in various ways, such as juxtaposition~\cite{chen2017sca-cnn:}, sequential~\cite{woo2018cbam:}, or integrated~\cite{hu2018gather-excite:}.

The studies above either aggregate the features to complement each other or enhance the diversity of the feature maps after a simple encoder. In contrast, the Inner-Imaging design considers both synchronously. We creatively use convolutional filters to organize channel signals on a pseudo-image, like putting lenses in the convolutional networks. This novel strategy reflects the cooperative grouping relations in multi-scale and achieves the integrated optimization of the diversity, complementarity, and completeness of CNN channels.

\section{Proposed Method}
\label{sec3}
In this section, the overall framework of the Inner-Imaging module is proposed, with a single type of G-filter or combined multi-shape G-filters. Subsequently, the special version of the InI module for ResNets is designed to jointly model the channel signals of residual flow and identity flow.

\subsection{Overall framework}
\label{sec3.1}
\begin{figure}[!t]
\centering
\includegraphics[scale=0.6]{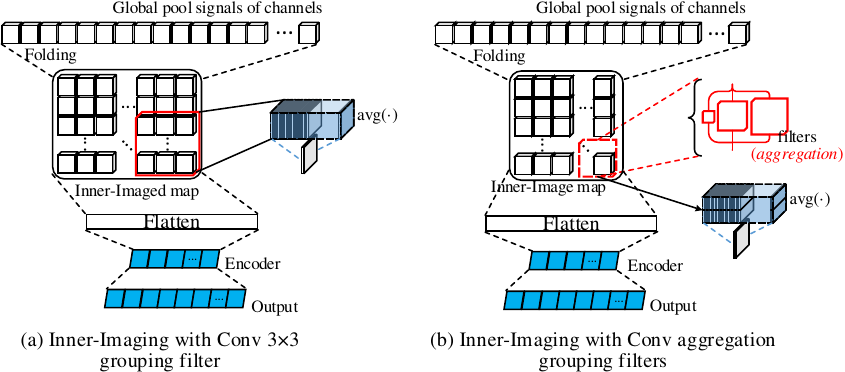}
\caption{The detailed structure of the Inner-Imaging. (a): The Inner-Imaging module with $3 \times 3$ convolutional G-filter; (b): The Inner-Imaging module with multi-shape G-filters aggregation.}
\label{fig2}
\end{figure}

In each layer of CNNs, each convolutional kernel produces a feature map $\mathrm{\textbf{u}}_{l}^{c} \in \mathbb{R}^{W \times H}$, which is defined as a channel. It forms the basic unit of intermediate features in convolutional networks. In order to model the relationships between them, the feature maps $\mathrm{\textbf{u}}_{l}^{c}$ are squeezed firstly, as follows:
\begin{equation}\label{eq1}
\mathrm{\textbf{u}}_{l}^{c} \in \mathrm{\textbf{U}}_{l} = {F}_l (\mathrm{\textbf{U}}_{l-1}, \mathrm{\textbf{W}}_{l}) = \left[ \mathrm{\textbf{u}}_{l}^{1}, \mathrm{\textbf{u}}_{l}^{2}, \dots, \mathrm{\textbf{u}}_{l}^{C} \right],
\end{equation}
\begin{equation}\label{eq2}
\hat{u}_{l}^{c} = {F}_{sq}(\mathrm{\textbf{u}}_{l}^{c}) = AvgGlobalPool(\mathrm{\textbf{u}}_{l}^{c}),
\end{equation}
\begin{equation}\label{eq3}
\hat{\mathrm{\textbf{v}}}_{init} = \left[ \hat{u}_{l}^{1}, \hat{u}_{l}^{2}, \dots, \hat{u}_{l}^{C} \right] \in \mathbb{R}^{1 \times C},
\end{equation}
where $\mathrm{\textbf{U}}_{l} = [\mathrm{\textbf{u}}_{l}^{1}, \mathrm{\textbf{u}}_{l}^{2}, \dots, \mathrm{\textbf{u}}_{l}^{C}]$ refers to the feature maps in the layer $l$ and $C$ is the number of channels, ${F}_{l}(\cdot)$ is the function of convolutional layer parameterized by $\mathrm{\textbf{W}}_{l}$, the function $AvgGlobalPool(\ast)$ is global average pooling, and ${F}_{sq} (\cdot)$ denotes the squeeze function with global average pooling. The channel signals $[\hat{u}_{l}^{1}, \hat{u}_{l}^{2}, \dots, \hat{u}_{l}^{C}]$ can be obtained from feature maps $\mathrm{\textbf{U}}_{l}$, and the initial channel signal map $\hat{\mathrm{\textbf{v}}}_{init}$ with the shape of $(1 \times C)$ is also constructed.

Next, by scanning the channel signal map with G-filters $\mathrm{\textbf{w}}^{(a \times b)}$, the channels in the same receptive field with the shape of $(a \times b)$ are organized as a group. However, most G-filters are not available on the initial map $\mathrm{\textbf{v}}_{init}$ unless $a=1$. Therefore, a new map, named as the inner-imaged map, is generated as:
\begin{equation}\label{eq4}
\begin{aligned}
\hat{\mathrm{\textbf{v}}}_{f} &= {T}(\hat{\mathrm{\textbf{v}}}_{init}) = {T} \left( \left[ \hat{u}_{l}^{1}, \hat{u}_{l}^{2}, \dots, \hat{u}_{l}^{C} \right] \right) \\
&= \left[
\begin{matrix}
  \hat{u}_{l}^{11} & \cdots & \hat{u}_{l}^{1M} \\
  \vdots & \ddots & \vdots \\
  \hat{u}_{l}^{N1} & \cdots & \hat{u}_{l}^{NM}
\end{matrix}
\right] \in \mathbb{R}^{N \times M}, (N \times M = C),
\end{aligned}
\end{equation}
where $\hat{\mathrm{\textbf{v}}}_{f}$ denotes the inner-imaged map which is folded by the reshape function ${T}(\cdot)$, its shape is $(N \times M)$. Then, the grouping relations between CNN channels are modeled as follows:
\begin{equation}\label{eq5}
\mathrm{\textbf{v}}' [i] = \hat{\mathrm{\textbf{v}}}_{f} * \mathrm{\textbf{w}}_{i}^{(a \times b)} = \left[
\begin{matrix}
  {v}_{i}^{11} & \cdots & {v}_{i}^{1m} \\
  \vdots & \ddots & \vdots \\
  {v}_{i}^{n1} & \cdots & {v}_{i}^{nm}
\end{matrix}
\right],
\end{equation}
\begin{equation}\label{eq6}
\bar{v}^{xy} = \left( \sum_{i=1}^{\varepsilon} \left( {v}_{i}^{xy} \right) \right) \Big/ \varepsilon
\end{equation}
\begin{equation}\label{eq7}
\bar{\mathrm{\textbf{v}}}' = \frac{1}{\varepsilon} \sum_{i=1}^{\varepsilon} \left( \mathrm{\textbf{v}}' [i] \right) = \left[
\begin{matrix}
  \bar{v}^{11} & \cdots & \bar{v}^{1m} \\
  \vdots & \ddots & \vdots \\
  \bar{v}^{n1} & \cdots & \bar{v}^{nm}
\end{matrix}
\right] \in \mathbb{R}^{n \times m}
\end{equation}
\begin{equation}\label{eq8}
\begin{aligned}
\bar{\mathrm{\textbf{v}}} &= \left( {F}_{flatten}(\bar{\mathrm{\textbf{v}}}') \right)^{\top} = {\left[ \bar{v}^{1}, \dots, \bar{v}^{{C}^{g}} \right]}^{\top} \in \mathbb{R}^{{C}^{g} \times 1} \\
&= {\left[ \bar{v}^{11}, \dots, \bar{v}^{1m}, \bar{v}^{21}, \dots, \bar{v}^{2m}, \dots, \bar{v}^{nm} \right]}^{\top}, \\
& \ (n \times m = {C}^{g})
\end{aligned}
\end{equation}

In Eq.~\ref{eq5}$-$\ref{eq8}, the operator $*$ denotes the convolution, and $\mathrm{\textbf{v}}' [i]$ refers to the convolutional result of G-filter $\mathrm{\textbf{w}}_{i}^{(a \times b)}$ on the inner-imaged map, and its shape is $(n \times m)$, $\varepsilon$ is the number of G-filters. The convolutional results are averaged and then applied to obtain the grouping map $\bar{\mathrm{\textbf{v}}}'$ whose element is $\bar{v}^{xy}$. Finally, the grouping map $\bar{\mathrm{\textbf{v}}}'$ is flattened as the tensor $\bar{\mathrm{\textbf{v}}} \in \mathbb{R}^{{C}^{g} \times 1}$, where each element of $\bar{\mathrm{\textbf{v}}}$ is a group signal modeled by $\mathrm{\textbf{w}}_{i}^{(a \times b)}$, and ${C}^{g}$ is the number of modeled channel groups.

Obviously, the diversified multi-shape G-filters can be integrated, such as $\{ \mathrm{\textbf{w}}^{(a_1 \times b_1)}, \mathrm{\textbf{w}}^{(a_2 \times b_2)}, \dots, \mathrm{\textbf{w}}^{(a_p \times b_p)} \}$, which can be applied to organize the channel groups with different sizes as follows:
\begin{equation}\label{eq9}
\mathrm{\textbf{v}}_{j}' [i] = \left( \hat{\mathrm{\textbf{v}}}_{f} * \mathrm{\textbf{w}}_{i}^{(a_j \times b_j)} \right) \in \mathbb{R}^{n_j \times m_j},
\end{equation}
\begin{equation}\label{eq10}
\mathrm{\textbf{v}}_{1:p}' [i] = \left[ \mathrm{\textbf{v}}_{1}' [i] \Join \mathrm{\textbf{v}}_{2}' [i] \Join \dots \Join \mathrm{\textbf{v}}_{p}' [i] \right],
\end{equation}
\begin{equation}\label{eq11}
\begin{aligned}
\bar{\mathrm{\textbf{v}}}_{1:p}' &= \left( \frac{1}{\varepsilon} \sum_{i=1}^{\varepsilon} \left( \mathrm{\textbf{v}}_{1:p}' [i] \right) \right) \in \mathbb{R}^{({n}_{1:p)} \times ({m}_{1:p})} \\
&, \quad {n}_{1:p} = \mathop{\mathrm{\textbf{\textbf{max}}}} \limits_{j=1} \limits^{p} \left( n_j \right), {m}_{1:p} = \sum \nolimits_{j=1} \nolimits^{p} \left( m_j \right),
\end{aligned}
\end{equation}
where $\mathrm{\textbf{v}}_{j}' [i]$ with the shape of $(n_j \times m_j)$ indicates the convolutional result of the G-filter $\mathrm{\textbf{w}}_{i}^{(a_j \times b_j)}$ and $p$ is the number of types of G-filter. $\mathrm{\textbf{v}}_{1:p}' [i]$ is the concatenated result of all types G-filters, and $\Join$ means matrix concatenation. Since $v'_ {1: p} [i]$ is a concatenation of $v'_1[i]$, $v'_2[i]$, ..., $v'_p[i]$, the number of columns $m_ {1: p}$ of $v'_{1:p}[i]$ is the sum of the number of columns $m_j$ of $v'_1[i]$, $v'_2[i]$, ..., $v'_p[i]$. Further, since $\bar{v}'_{1: p}$ is the result of matrix summing of all $v'_{1: p}[i]$, they share the same number of columns $m_{1: p}$. In the case of different shapes of the matrix, zero is automatically filled into the blank. Then, $\bar{\mathrm{\textbf{v}}}_{1:p}'$ is the average result of each $\varepsilon$ G-filters with different shapes and sizes. $(({n}_{1:p)} \times ({m}_{1:p}))$ is the final shape of the averaged group signal map.

The grouping map is still flatten as:
\begin{equation}\label{eq12}
\begin{aligned}
\bar{\mathrm{\textbf{v}}} &= \left( {F}_{flatten}(\bar{\mathrm{\textbf{v}}}_{1:p}') \right)^{\top} = {\left[ \bar{v}^{1}, \dots, \bar{v}^{{C}^{g}} \right]}^{\top} \in \mathbb{R}^{{C}^{g} \times 1}, \\
& \ \left( ({n}_{1:p)} \times ({m}_{1:p}) = {C}^{g} \right)
\end{aligned}
\end{equation}
thus, a greater ${C}^{g}$ than that in Eq.~\ref{eq8} is obtained, leading to more samples for the group relationships of CNN channels.

After representing the channel groups as $\bar{\mathrm{\textbf{v}}} \in \mathbb{R}^{{C}^{g} \times 1}$, the relations are encoded between group signals with FC layers $\mathrm{\textbf{w}}^{1} \in \mathbb{R}^{\frac{C}{t} \times {C}^{g}}$ and $\mathrm{\textbf{w}}^{2} \in \mathbb{R}^{ {C} \times \frac{C}{t}}$. Then, channel-wise attention is conducted as follows:
\begin{equation}\label{eq13}
\mathrm{\textbf{W}}_{att} = \mathop{\cup} \limits_{j=1} \limits^{p} \{\mathrm{\textbf{w}}_{1}^{(a_j \times b_j)}, \dots, \mathrm{\textbf{w}}_{\varepsilon}^{(a_j \times b_j)} \} \cup \{ \mathrm{\textbf{w}}^{1}, \mathrm{\textbf{w}}^{2} \},
\end{equation}
\begin{equation}\label{eq14}
\mathrm{\textbf{s}} = {F}_{att} \left( \mathrm{\textbf{U}}_{l}, \mathrm{\textbf{W}}_{att} \right) = \sigma \left( \mathrm{\textbf{w}}^{2} \cdot ReLU \left( \mathrm{\textbf{w}}^{1} \cdot \bar{\mathrm{\textbf{v}}} \right) \right)
\end{equation}
\begin{equation}\label{eq15}
\mathrm{\textbf{U}}_{l}^{att} = \mathrm{\textbf{s}} \circ \mathrm{\textbf{U}}_{l} = {F}_{att} \left( \mathrm{\textbf{U}}_{l}, \mathrm{\textbf{W}}_{att} \right) \circ {F}_l (\mathrm{\textbf{U}}_{l-1}, \mathrm{\textbf{W}}_{l}),
\end{equation}
where $\mathrm{\textbf{W}}_{att}$ is the set of parameters in the InI module, which contains all parameters of the G-filters and the FC encoders, $\mathrm{\textbf{s}}$ is the channel-wise attentional outputs, ${F}_{att}$ denotes the summarized function of channel-wise attention, operator $\circ$ is the elements-wise, product and $\mathrm{\textbf{U}}_{l}^{att}$ refers to the attentional feature maps. Like all conventional channel-wise attention structures, $\mathrm{\textbf{U}}_{l}^{att}$ refers to the final feature map, which is the product of the original feature map and the attentional value. The number of G-filters $\varepsilon$ is set to $C/t$, where $t$ is the dimensionality-reduction ratio, it is set to 16 as~\cite{hu2018squeeze-and-excitation}.

Fig.~\ref{fig2} shows the detailed structure of the InI module.

\subsection{Special version for ResNets}
\label{sec3.2}
\begin{figure}[!t]
\centering
\includegraphics[scale=0.65]{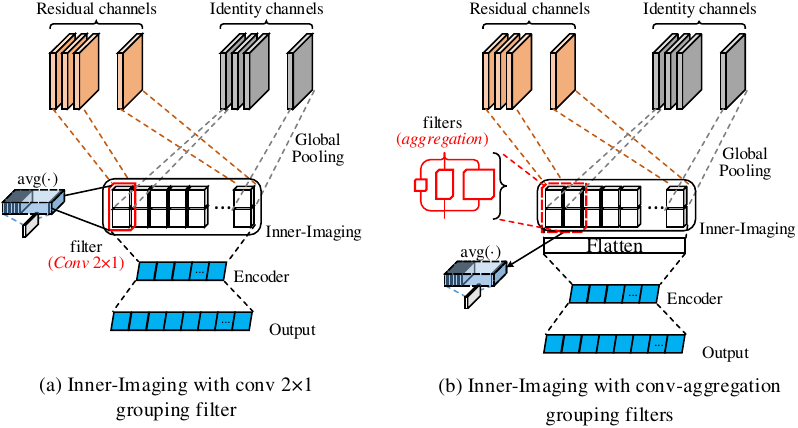}
\caption{The special version of simplified Inner-Imaging for ResNets. (a): The simplified Inner-Imaging module with $2 \times 1$ convolutional G-filter; (b): The simplified Inner-Imaging module with multi-shape G-filters aggregation.}
\label{fig3}
\end{figure}

In ResNets, the residual flow is considered as a complement to the identity mapping~\cite{he2016identity}, inspired by this argument. We propose the special version of the InI mechanism for ResNets. It attempts to expand the scope of channel relation modeling to both residual and identity channels. This strategy helps residual mappings to supplement identity mappings more precisely.

We define the identity and residual mappings as $\mathrm{\textbf{X}}_{l}$ and $\mathrm{\textbf{U}}_{l}$, respectively. The feature map $\mathrm{\textbf{x}}_{l}^{c}$ in the identity mappings can be pooled in a similar way to Eq.~\ref{eq1}. As shown in Fig.~\ref{fig3}, the channel signals of $\mathrm{\textbf{X}}_{l}$ and $\mathrm{\textbf{U}}_{l}$ can be simply stacked without the operation of folding, as follows:
\begin{equation}\label{eq16}
\mathrm{\textbf{x}}_{l}^{c} \in \mathrm{\textbf{X}}_{l} = \left[ \mathrm{\textbf{x}}_{l}^{1}, \mathrm{\textbf{x}}_{l}^{2}, \dots, \mathrm{\textbf{x}}_{l}^{C} \right],
\end{equation}
\begin{equation}\label{eq17}
\hat{x}_{l}^{c} = {F}_{sq}(\mathrm{\textbf{x}}_{l}^{c}) = AvgGlobalPool(\mathrm{\textbf{x}}_{l}^{c}),
\end{equation}
\begin{equation}\label{eq18}
\hat{\mathrm{\textbf{x}}}_{l} = \left[ \hat{x}_{l}^{1}, \hat{x}_{l}^{2}, \dots, \hat{x}_{l}^{C} \right] \in \mathbb{R}^{1 \times C},
\end{equation}
\begin{equation}\label{eq19}
\hat{\mathrm{\textbf{v}}}_{stack} = \left[
  \begin{matrix}
   \hat{\mathrm{\textbf{u}}}_{l} \\
   \hat{\mathrm{\textbf{x}}}_{l}
  \end{matrix}
  \right] = \left[
 \begin{matrix}
   \hat{u}_{l}^{1} & \hat{u}_{l}^{2} & \cdots & \hat{u}_{l}^{C} \\
   \hat{x}_{l}^{1} & \hat{x}_{l}^{2} & \cdots & \hat{x}_{l}^{C}
  \end{matrix}
  \right] \in \mathbb{R}^{2 \times C},
\end{equation}
\begin{equation}\label{eq20}
\bar{\mathrm{\textbf{v}}}' = \frac{1}{\varepsilon} \sum_{i=1}^{\varepsilon} \left( \hat{\mathrm{\textbf{v}}}_{stack} * \mathrm{\textbf{w}}_{i}^{(a \times b)} \right), \ (a \leqslant 2),
\end{equation}
\begin{equation}\label{eq21}
\bar{\mathrm{\textbf{v}}} = \left\{
\begin{array}{lr}
\left( \bar{\mathrm{\textbf{v}}}' \right)^{\top}, & \mathrm{if} \ {a = 2} \\
\left( {F}_{flatten} \left( \bar{\mathrm{\textbf{v}}}' \right) \right)^{\top}, & \mathrm{if} \ {a < 2}
\end{array}
\right\} \in \mathbb{R}^{{C}^{g} \times 1}
\end{equation}
where $\hat{x}_{l}^{c}$ is the squeezed signal of identity feature maps ${x}_{l}^{c}$, $\hat{\mathrm{\textbf{v}}}_{stack}$ is the simplified inner-imaged map by stacking the channel signals $\hat{\mathrm{\textbf{u}}}_{l}$ and $\hat{\mathrm{\textbf{x}}}_{l}$. Multi-shape G-filters can also be applied to $\hat{\mathrm{\textbf{v}}}_{stack}$, as follows:
\begin{equation}\label{eq22}
\begin{aligned}
\bar{\mathrm{\textbf{v}}}' = \frac{1}{\varepsilon} \sum_{i=1}^{\varepsilon} \left( {F}_{norm} \left( \hat{\mathrm{\textbf{v}}}_{stack} * \mathrm{\textbf{w}}_{i}^{({a}_{1} \times {b}_{1})}, \dots \right. \right. \\
\left. \left. , \hat{\mathrm{\textbf{v}}}_{stack} * \mathrm{\textbf{w}}_{i}^{({a}_{p} \times {b}_{p})} \right) \right), \ (\forall {a}_{j}: {a}_{j} \leqslant 2),
\end{aligned}
\end{equation}
\begin{equation}\label{eq23}
\bar{\mathrm{\textbf{v}}} = \left\{
\begin{array}{lr}
\left( \bar{\mathrm{\textbf{v}}}' \right)^{\top}, & \mathrm{if} \ { \forall {a}_{j}: {a}_{j} = 2} \\
\left( {F}_{flatten} \left( \bar{\mathrm{\textbf{v}}}' \right) \right)^{\top}, & \mathrm{if} \ { \exists {a}_{j}: {a}_{j} < 2}
\end{array}
\right.
\end{equation}

Although the step of folding the original channel signal map is omitted here, it has a considerable limitation on the shape of group filters: $\forall {a}: {a} \leqslant 2$. So, we fold $\hat{\mathrm{\textbf{v}}}_{stack}$ as:
\begin{equation}\label{eq24}
\begin{aligned}
\hat{\mathrm{\textbf{v}}}_{f} &= {T}_{alt} \left(\hat{\mathrm{\textbf{v}}}_{stack} \right) = \left[
\begin{matrix}
  \hat{v}_{f}^{11} & \cdots & \hat{v}_{f}^{1M} \\
  \vdots & \ddots & \vdots \\
  \hat{v}_{f}^{N1} & \cdots & \hat{v}_{f}^{NM}
\end{matrix}
\right] \\
&= {T}_{alt} \left(\left[
  \begin{matrix}
   \hat{\mathrm{\textbf{x}}}_{l} \\
   \hat{\mathrm{\textbf{u}}}_{l}
  \end{matrix}
  \right] \right) = \left[
  \begin{matrix}
    \hat{x}_{l}^{1} & \hat{x}_{l}^{2} & \cdots & \hat{x}_{l}^{m} \\
    \hat{u}_{l}^{1} & \hat{u}_{l}^{2} & \cdots & \hat{u}_{l}^{m} \\
    \vdots & \vdots & \ddots & \vdots \\
    \hat{x}_{l}^{\frac{N}{2}1} & \hat{x}_{l}^{\frac{N}{2}2} & \cdots & \hat{x}_{l}^{\frac{N}{2}M} \\
    \hat{u}_{l}^{\frac{N}{2}1} & \hat{u}_{l}^{\frac{N}{2}2} & \cdots & \hat{u}_{l}^{\frac{N}{2}M}
  \end{matrix}
  \right] \\
& \in \mathbb{R}^{N \times M}, (N \times M = 2C),
\end{aligned}
\end{equation}
where $\hat{\mathrm{\textbf{v}}}_{f}$ is the folded inner-imaged map with $\hat{\mathrm{\textbf{x}}}_{l}$ and $\hat{\mathrm{\textbf{u}}}_{l}$, ${T}_{alt}$ is the alternating reshape function, which enables both residual and identity signals to be scanned in one receptive field.

\begin{figure}[!t]
\centering
\includegraphics[scale=0.7]{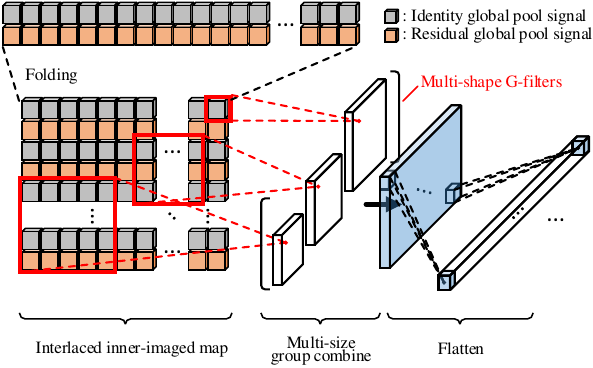}
\caption{The folded inner-imaged map of ResNets and the subsequent modeling with multi-shape G-filters.}
\label{fig4}
\end{figure}

Fig.~\ref{fig4} shows the structure of the special version InI module on the folded inner-imaged map, with multi-shape G-filters aggregation. The subsequent process follows Eq.~\ref{eq9}$-$\ref{eq11}; the only difference is that the grouping relations of both residual and identity channels are modeled.

FC encoders $\mathrm{\textbf{w}}^{1} \in \mathbb{R}^{\frac{C}{t} \times {C}^{g}}$ and $\mathrm{\textbf{w}}^{2} \in \mathbb{R}^{ {C} \times \frac{C}{t}}$ are still used to model the group-wise relations and output the final channel-wise attentional values. As the definition of residual unit~\cite{he2016deep}:
\begin{equation}\label{eq25}
y = \mathrm{\textbf{X}}_{l} + \mathrm{\textbf{U}}_{l} = \mathrm{\textbf{X}}_{l} + {F}_{res} \left( \mathrm{\textbf{X}}_{l}, \mathrm{\textbf{W}}_{l} \right),
\end{equation}
we obtain:
\begin{equation}\label{eq26}
\mathrm{\textbf{s}} = {F}_{att} \left( \left( \mathrm{\textbf{X}}_{l}, \mathrm{\textbf{U}}_{l} \right), \mathrm{\textbf{W}}_{att} \right),
\end{equation}
\begin{equation}\label{eq27}
\begin{aligned}
y &= \mathrm{\textbf{X}}_{l} + \mathrm{\textbf{s}} \circ \mathrm{\textbf{U}}_{l} \\
&= \mathrm{\textbf{X}}_{l} + {F}_{att} \left( \left( \mathrm{\textbf{X}}_{l}, \mathrm{\textbf{U}}_{l} \right), \mathrm{\textbf{W}}_{att} \right) \circ {F}_{res} \left( \mathrm{\textbf{X}}_{l}, \mathrm{\textbf{W}}_{l} \right),
\end{aligned}
\end{equation}
where $\mathrm{\textbf{s}}$ refers to the outputs of channel-wise attention, and $\mathrm{\textbf{W}}_{att}$ is the total set of parameters in the InI module, which is defined by Eq.~\ref{eq13}.

\section{Theories}
\label{sec4}
In this section, some theoretical details of the InI model are elaborated and discussed, including its technical advantages and some insightful design motivations.

\subsection{The insight of Inner-Imaging mechanism}
\label{sec4.1}
It is an elegant design method to give new functions and meanings to existing tools. The convolutional filter is usually used to model the spatial features of vision data. Also, it can help us to model the grouping relations of convolutional channels very conveniently.

Compare with the typical channel-wise attention mechanism. The InI module can provide more exhaustive and diverse modeling of channel relations, especially on grouping relations.

In this section, we analyze the detailed operation of the InI module and compare it with the conventional channel-wise attention. To complete this comparison, we first review the process of original channel-wise attention: excluding the output layer $\mathrm{\textbf{w}}^{2}$, we discuss the FC encoder $\mathrm{\textbf{w}}^{1}$, as follows:
\begin{equation}\label{eq28}
\mathrm{\textbf{w}}^{1} = \left[
\begin{matrix}
  {w}_{1}^{11} & \cdots & {w}_{1}^{1 {C}} \\
  \vdots & \ddots & \vdots \\
  {w}_{1}^{\frac{C}{t} 1} & \cdots & {w}_{1}^{\frac{C}{t} {C}}
\end{matrix}
\right] \in \mathbb{R}^{\frac{C}{t} \times {C}},
\end{equation}
\begin{equation}\label{eq29}
\begin{aligned}
\mathrm{\textbf{e}} &= \mathrm{\textbf{w}}^{1} \cdot \left( \hat{\mathrm{\textbf{v}}}_{init} \right)^{\top} = \left[
\begin{matrix}
  {w}_{1}^{11} & \cdots & {w}_{1}^{1 {C}} \\
  \vdots & \ddots & \vdots \\
  {w}_{1}^{\frac{C}{t} 1} & \cdots & {w}_{1}^{\frac{C}{t} {C}}
\end{matrix}
\right] \cdot \left[
\begin{matrix}
  \hat{u}_{l}^{1} \\
  \vdots \\
  \hat{u}_{l}^{C}
\end{matrix}
\right] \\
&= \left[ \sum \nolimits_{i=1} \nolimits^{C} {w}_{1}^{1 i} \hat{u}_{l}^{i}, \dots, \sum \nolimits_{i=1} \nolimits^{C} {w}_{1}^{\frac{C}{t} i} \hat{u}_{l}^{i} \right] \\
&= {\left[e \left(\hat{\mathrm{\textbf{v}}}_{init}, \mathrm{\textbf{w}}_{1}^{1 [\cdot]} \right), \dots, e \left(\hat{\mathrm{\textbf{v}}}_{init}, \mathrm{\textbf{w}}_{1}^{\frac{C}{t} [\cdot]} \right) \right]} \in \mathbb{R}^{1 \times \frac{C}{t}},
\end{aligned}
\end{equation}
where $\mathrm{\textbf{e}}$ denotes the embedding of CNN channels, $e ( \cdot, \mathrm{\textbf{w}}_{1}^{i [\cdot]} )$ indicates the feature parameterized by weights $\mathrm{\textbf{w}}_{1}^{i [\cdot]}$.

In InI mechanism, there are two successive stages: (a) grouping modeling; (b) FC embedding, as follows:
\begin{equation}\label{eq30}
\begin{aligned}
\mathrm{\textbf{s}} &= {F}_{att} \Big( \mathrm{\textbf{U}}_{l}, \underbrace{ \{ \mathrm{\textbf{w}}^{(a_1 \times b_1)}, \dots, \mathrm{\textbf{w}}^{(a_p \times b_p)} \} }_{a}, \underbrace{ \{ \mathrm{\textbf{w}}^{1}, \mathrm{\textbf{w}}^{2} \} }_{b} \Big), \\
& \ \Big(\mathrm{\textbf{w}}^{(a_j \times b_j)} = \big[\mathrm{\textbf{w}}_{1}^{(a_j \times b_j)}, \dots, \mathrm{\textbf{w}}_{\varepsilon}^{(a_j \times b_j)} \big] \Big),
\end{aligned}
\end{equation}

When the convolutions $\mathrm{\textbf{w}}_{1}^{(a_j \times b_j)}, \dots, \mathrm{\textbf{w}}_{\varepsilon}^{(a_j \times b_j)}$ are abbreviated as:
\begin{equation}\label{eq31}
\sum_{i=1}^{\varepsilon} \left( \hat{\mathrm{\textbf{v}}}_{f} * \mathrm{\textbf{w}}_{i}^{(a_j \times b_j)} \right) \Rightarrow \hat{\mathrm{\textbf{v}}}_{f} * \mathrm{\textbf{w}}^{(a_j \times b_j)},
\end{equation}
we obtain:
\begin{equation}\label{eq32}
\bar{\mathrm{\textbf{v}}}' = \hat{\mathrm{\textbf{v}}}_{f} * \mathrm{\textbf{w}}^{(a_1 \times b_1)} \Join \cdots \Join \hat{\mathrm{\textbf{v}}}_{f} * \mathrm{\textbf{w}}^{(a_p \times b_p)}
\end{equation}
\begin{equation}\label{eq33}
\begin{aligned}
\hat{\mathrm{\textbf{v}}}_{f} * \mathrm{\textbf{w}}^{(a_j \times b_j)} &= \left[
\begin{matrix}
  {v}^{1 1} & \cdots & {v}^{1 m_j} \\
  \vdots & \ddots & \vdots \\
  {v}^{n_j 1} & \cdots & {v}^{n_j m_j}
\end{matrix}
\right] \in \mathbb{R}^{n_j \times m_j} \\
&= \left[
\begin{matrix}
  {e}_{g}^{j}(\hat{\mathrm{\textbf{g}}}_{j}^{1}) & \cdots & {e}_{g}^{j}(\hat{\mathrm{\textbf{g}}}_{j}^{m_j}) \\
  \vdots & \ddots & \vdots \\
  {e}_{g}^{j}(\hat{\mathrm{\textbf{g}}}_{j}^{{C}_{g}^{j} - m_j + 1}) & \cdots & {e}_{g}^{j}(\hat{\mathrm{\textbf{g}}}_{j}^{{C}_{g}^{j}})
\end{matrix}
\right]
\end{aligned}
\end{equation}
\begin{equation}\label{eq34}
\begin{aligned}
\mathrm{\textbf{w}}^{(a_j \times b_j)} &= \left[ \big[ {w}_{j}^{11}, \dots, {w}_{j}^{1 b_j} \big], \dots, \big[ {w}_{j}^{a_j 1}, \dots, {w}_{j}^{a_j b_j} \big] \right] \\
& \Rightarrow \left[ {w}_{j}^{1}, \dots, {w}_{j}^{{\theta}_j} \right], \ ( a_j \times b_j = {\theta}_j ),
\end{aligned}
\end{equation}
\begin{equation}\label{eq35}
\begin{aligned}
\hat{\mathrm{\textbf{g}}}_{j}^{k} &= \left[ \big[ \mathop{\hat{v}^{11}} \limits_{\to k}, \dots, \mathop{\hat{v}^{1 b_j}} \limits_{\to k} \big], \dots, \big[ \mathop{\hat{v}^{a_j 1}} \limits_{\to k}, \dots, \mathop{\hat{v}^{a_j b_j}} \limits_{\to k} \big] \right] \\
& \Rightarrow \left[ \hat{v}_{k}^{1}, \hat{v}_{k}^{2}, \dots, \hat{v}_{k}^{{\theta}_j} \right]
\end{aligned}
\end{equation}
\begin{equation}\label{eq36}
\begin{aligned}
{e}_{g}^{j} \left(\hat{\mathrm{\textbf{g}}}_{j}^{k} \right) & \Leftarrow {e}_{g} \left(\hat{\mathrm{\textbf{g}}}_{j}^{k}, \mathrm{\textbf{w}}^{(a_j \times b_j)} \right) \\
&= \sum_{x_k = 1}^{a_j} \sum_{y_k = 1}^{b_j} {w}_{j}^{x_k y_k} \hat{v}_{f}^{x_k y_k} = \sum_{i=1}^{{\theta}_j} {w}_{j}^{i} \hat{v}_{k}^{i}
\end{aligned}
\end{equation}
where $\theta_{j}$ is the number of parameters in the G-filter $\mathrm{\textbf{w}}^{(a_j \times b_j)}$. $\hat{\mathrm{\textbf{g}}}_{j}^{k}$ is the $k$-th receptive field of G-filter $\mathrm{\textbf{w}}^{(a_j \times b_j)}$, ${\to k}$ means that the G-filter slides to the $k$-th receptive field. ${e}_{g}^{j} ( \cdot )$ is the encoded feature of each channel group.

Retrospect Eq.~\ref{eq29}, it is noticed that the FC encoder is a particular form in Eq.~\ref{eq36}, it has only one group which contains all channels.

Next, the role of the FC layer $\mathrm{\textbf{w}}^{1}$ changes, as follows:
\begin{equation}\label{eq37}
\begin{aligned}
\mathrm{\textbf{e}} &= \mathrm{\textbf{w}}^{1} \cdot \bar{\mathrm{\textbf{v}}} = \mathrm{\textbf{w}}^{1} \cdot \left( {F}_{flatten} \left( \bar{\mathrm{\textbf{v}}}' \right) \right)^{\top} \\
&= \left[ \sum_{k=1}^{{C}_{g}} {w}_{1}^{1 k} {e}_{g} \Big( \hat{\mathrm{\textbf{g}}}^{k} \Big), \dots, \sum_{k=1}^{{C}_{g}} {w}_{1}^{\frac{C}{t} k} {e}_{g} \Big( \hat{\mathrm{\textbf{g}}}^{k} \Big) \right],
\end{aligned}
\end{equation}
\begin{equation}\label{eq38}
\Gamma \left( \hat{\mathrm{\textbf{g}}}_{j}^{k} \right) \in \left\{ (a_j, b_j) : j = 1, 2, \dots , p \right\},
\end{equation}
\begin{equation}\label{eq39}
\begin{aligned}
\bar{\mathrm{\textbf{v}}} &= \Big[ {e}_{g}(\hat{\mathrm{\textbf{g}}}^{1}), \dots {e}_{g}(\hat{\mathrm{\textbf{g}}}^{{C}_{g}}) \Big] \\
&= \Big[ {e}_{g}^{1}(\hat{\mathrm{\textbf{g}}}_{1}^{1}), \dots {e}_{g}^{1}(\hat{\mathrm{\textbf{g}}}_{1}^{{C}_{g}^{1}}) \Big] \Join \cdots \\
& \quad  \Join \Big[ {e}_{g}^{p}(\hat{\mathrm{\textbf{g}}}_{p}^{1}), \dots {e}_{g}^{p}(\hat{\mathrm{\textbf{g}}}_{p}^{{C}_{g}^{p}}) \Big], \ \left( {C}_{g} = \sum \nolimits_{j=1}^{p} {C}_{g}^{j} \right),
\end{aligned}
\end{equation}
where $\Gamma(\cdot)$ denotes the shape of a matrix, and the one-hot G-filter $\mathrm{\textbf{o}} = [\alpha] \in \mathbb{R}^{1 \times 1}$ plays an important role. We obtain:
\begin{equation}\label{eq40}
\begin{aligned}
\hat{\mathrm{\textbf{v}}}_{f} * \mathrm{\textbf{o}} = \left[
\begin{matrix}
  {e}_{g}(\hat{\mathrm{\textbf{g}}}_{o}^{1}) & \cdots & {e}_{g}(\hat{\mathrm{\textbf{g}}}_{o}^{M}) \\
  \vdots & \ddots & \vdots \\
  {e}_{g}(\hat{\mathrm{\textbf{g}}}_{o}^{{C} - M + 1}) & \cdots & {e}_{g}(\hat{\mathrm{\textbf{g}}}_{o}^{C})
\end{matrix}
\right] \\
\in \mathbb{R}^{N \times M}, (N \times M = C),
\end{aligned}
\end{equation}
\begin{equation}\label{eq41}
\hat{\mathrm{\textbf{g}}}_{o}^{k} = \big[ \mathop{\hat{v}^{11}} \limits_{\to k} \big] = \left[ \hat{v}_{f}^{xy} \right], (x \times y = k),
\end{equation}
\begin{equation}\label{eq42}
{e}_{g} \left(\hat{\mathrm{\textbf{g}}}_{o}^{k} \right) = \alpha \cdot \hat{v}_{f}^{xy} \propto \hat{u}_{l}^{k} \in \hat{\mathrm{\textbf{v}}}_{init},
\end{equation}
where $\propto$ means proportional relationship, and ${e}_{g} \left(\hat{\mathrm{\textbf{g}}}_{o}^{k} \right)$ indicates the case of which single channel constructs a group. So,
\begin{equation}\label{eq43}
\begin{aligned}
\sum_{k=1}^{{C}_{g}} {w}_{1}^{i k} {e}_{g} \Big( \hat{\mathrm{\textbf{g}}}^{k} \Big) &= \underbrace{ \sum_{k=1}^{C} {w}_{1}^{i k} {e}_{g} \Big( \hat{\mathrm{\textbf{g}}}_{o}^{k} \Big) }_{channels} + \underbrace{ \sum_{k=1}^{{C}_{\neg o}} {w}_{1}^{i k} {e}_{g} \Big( \hat{\mathrm{\textbf{g}}}_{\neg o}^{k} \Big) }_{groups} \\
&, \ ({C}_{g} - C = {C}_{\neg o}),
\end{aligned}
\end{equation}
where $\neg o$ indicates the non-one-hot G-filter $\mathrm{\textbf{o}}$.

It can be seen that in the two stages of the InI mechanism, \textbf{stage-\textit{a}:} the G-filters generate channel groups in diversified shapes and model channel relations within them, as Eq.~\ref{eq32}$-$\ref{eq36}; \textbf{stage-\textit{b}:} the function of FC encoder $\mathrm{\textbf{w}}_{1}$ is applied to modeling the relations between various channel groups, as Eq.~\ref{eq37}$-$\ref{eq39}. The design of one-hot G-filter $\mathrm{\textbf{o}}$ adds the consideration of modeling the relationship between individual channels and channel groups, as Eq.~\ref{eq40}$-$\ref{eq43}.

It is concluded that the modeled channel relations by the InI mechanism include and much more abundant than that of the typical strategy.

\subsection{Joint modeling of residual and identity mappings}
\label{sec4.2}
\begin{figure*}[!t]
\centering
\includegraphics[scale=0.68]{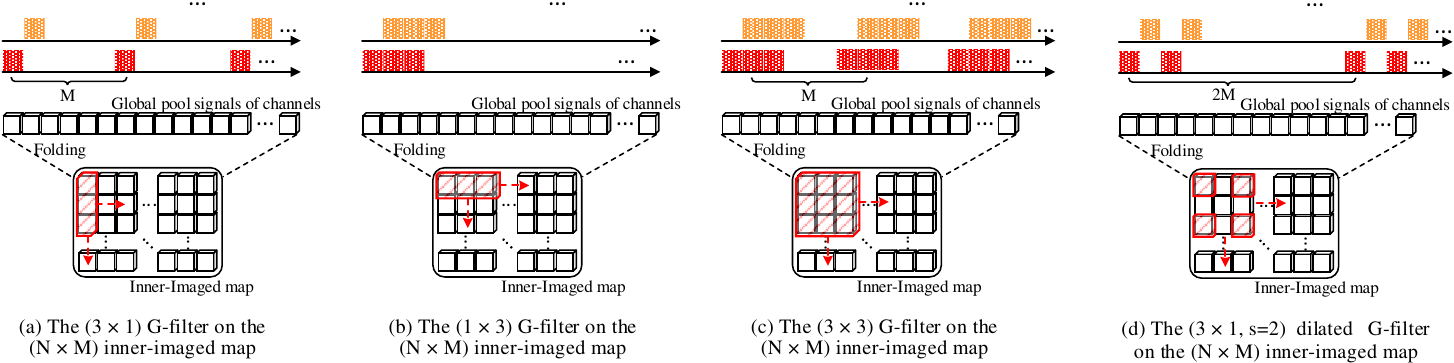}
\caption{Examples of channel grouping organization for multiple types of G-filters. (a): The G-filter with a vertical shape; (b): The G-filter with the horizontal shape with a horizontal shape; (c): The square G-filter; (d): The dilated G-filter.}
\label{fig5}
\end{figure*}

Another trick is proposed for the InI mechanism in ResNets, which is the joint modeling of residual and identity mappings. It is believed that this trick makes the ResNets more efficient.

As described in \cite{he2016identity}, the loss $\zeta$ is back-propagated (BP) as:
\begin{equation}\label{eq44}
\begin{aligned}
\frac{ \partial \zeta }{ \partial \mathrm{\textbf{X}}_{l} } &= \frac{ \partial \zeta }{ \partial \mathrm{\textbf{X}}_{L} } \frac{ \partial \mathrm{\textbf{X}}_{L} }{ \partial \mathrm{\textbf{X}}_{l} } \\
&= \frac{ \partial \zeta }{ \partial \mathrm{\textbf{X}}_{L} } \Big( 1 + \frac{ \partial }{ \partial \mathrm{\textbf{X}}_{l} } \sum_{i=l}^{L-1} {F}_{res} ( \mathrm{\textbf{X}}_{i}, \mathrm{\textbf{W}}_{l} ) \Big),
\end{aligned}
\end{equation}
where $L$ indicates any deeper residual unit, $l$ indicates any shallower unit, and after using channel-wise attention, we obtain:
\begin{equation}\label{eq45}
\begin{aligned}
\frac{ \partial \zeta }{ \partial \mathrm{\textbf{X}}_{l} } &= \frac{ \partial \zeta }{ \partial \mathrm{\textbf{X}}_{L} } \Big( 1 + \frac{ \partial }{ \partial \mathrm{\textbf{X}}_{l} } \sum_{i=l}^{L-1} {F}_{att} ( \mathrm{\textbf{U}}_{i} ) \\
& \qquad \qquad \qquad \circ {F}_{res} ( \mathrm{\textbf{X}}_{i}, \mathrm{\textbf{W}}_{i} ) \Big) \\
&= \frac{ \partial \zeta }{ \partial \mathrm{\textbf{X}}_{L} } \bigg( 1 + \sum_{i=l}^{L-1} \Big( {F}_{res} ( \mathrm{\textbf{X}}_{i}, \mathrm{\textbf{W}}_{i} ) \frac{ \partial {F}_{att} (\mathrm{\textbf{U}}_{i}) }{ \partial \mathrm{\textbf{X}}_{l} } \\
& \qquad \qquad \qquad + {F}_{att} (\mathrm{\textbf{U}}_{i}) \frac{ \partial {F}_{res} ( \mathrm{\textbf{X}}_{i}, \mathrm{\textbf{W}}_{i} ) }{ \partial \mathrm{\textbf{X}}_{l} } \Big)  \bigg), \\
\end{aligned}
\end{equation}
because $\mathrm{\textbf{U}}_{i} = {F}_{res} (\mathrm{\textbf{X}}_{i}, \mathrm{\textbf{W}}_{i})$, we obtain:
\begin{equation}\label{eq46}
\begin{aligned}
\frac{ \partial \zeta }{ \partial \mathrm{\textbf{X}}_{l} } &= \frac{ \partial \zeta }{ \partial \mathrm{\textbf{X}}_{L} } \bigg( 1 + \sum_{i=l}^{L-1} \Big( \mathrm{\textbf{U}}_{i} \frac{ \partial {F}_{att} ( \mathrm{\textbf{U}}_{i} ) }{ \partial \mathrm{\textbf{X}}_{l} } \\
& \qquad \qquad \qquad + {F}_{att} ( \mathrm{\textbf{U}}_{i} ) \frac{ \partial \mathrm{\textbf{U}}_{i} }{ \partial \mathrm{\textbf{X}}_{l} } \Big) \bigg) \\
&= \frac{ \partial \zeta }{ \partial \mathrm{\textbf{X}}_{L} } \bigg( 1 + \sum_{i=l}^{L-1} \Big( \mathrm{\textbf{U}}_{i} \frac{ \partial {F}_{att} ( \mathrm{\textbf{U}}_{i} ) }{ \partial \mathrm{\textbf{U}}_{i} } \frac{ \partial \mathrm{\textbf{U}}_{i} }{ \partial \mathrm{\textbf{X}}_{l} } \\
& \qquad \qquad \qquad + {F}_{att} ( \mathrm{\textbf{U}}_{i} ) \frac{ \partial \mathrm{\textbf{U}}_{i} }{ \partial \mathrm{\textbf{X}}_{l} } \Big) \bigg),
\end{aligned}
\end{equation}
where
$$
\beta = \mathrm{\textbf{U}}_{i} \frac{ \partial {F}_{att}( \mathrm{\textbf{U}}_{i} ) }{ \partial \mathrm{\textbf{U}}_{i} } + {F}_{att}( \mathrm{\textbf{U}}_{i} ),
$$
we obtain:
\begin{equation}\label{eq47}
\frac{ \partial \zeta }{ \partial \mathrm{\textbf{X}}_{l} } = \frac{ \partial \zeta }{ \partial \mathrm{\textbf{X}}_{L} } \bigg( 1 + \sum_{i=l}^{L-1} \Big( \beta \cdot \frac{ \partial \mathrm{\textbf{U}}_{i} }{ \partial \mathrm{\textbf{X}}_{l} } \Big) \bigg).
\end{equation}
Then, after we set
$$
\beta = \mathrm{\textbf{U}}_{i} \frac{ \partial {F}_{att}( \mathrm{\textbf{X}}_{i} \mathrm{\textbf{U}}_{i} ) }{ \partial \mathrm{\textbf{U}}_{i} } + {F}_{att}( \mathrm{\textbf{U}}_{i} ),
$$
and
$$
\gamma = \mathrm{\textbf{U}}_{i} \frac{ \partial {F}_{att}( \mathrm{\textbf{X}}_{i}, \mathrm{\textbf{U}}_{i} ) }{ \partial \mathrm{\textbf{X}}_{i} },
$$
with our strategy, we can obtain:
\begin{equation}\label{eq48}
\begin{aligned}
\frac{ \partial \zeta }{ \partial \mathrm{\textbf{X}}_{l} } &= \frac{ \partial \zeta }{ \partial \mathrm{\textbf{X}}_{L} } \Big( 1 + \frac{ \partial }{ \partial \mathrm{\textbf{X}}_{l} } \sum_{i=l}^{L-1} {F}_{att}( \mathrm{\textbf{X}}_{i}, \mathrm{\textbf{U}}_{i} ) \\
& \qquad \qquad \qquad \circ {F}_{res} ( \mathrm{\textbf{X}}_{i}, \mathrm{\textbf{W}}_{i} ) \Big) \\
&= \frac{ \partial \zeta }{ \partial \mathrm{\textbf{X}}_{L} } \bigg( 1 + \sum_{i=l}^{L-1} \Big( {F}_{res} ( \mathrm{\textbf{X}}_{i}, \mathrm{\textbf{W}}_{i} ) \frac{ \partial {F}_{att}( \mathrm{\textbf{X}}_{i}, \mathrm{\textbf{U}}_{i} ) }{ \partial \mathrm{\textbf{X}}_{l} } \\
& \qquad \qquad \qquad + {F}_{att}( \mathrm{\textbf{X}}_{i}, \mathrm{\textbf{U}}_{i} ) \frac{ \partial {F}_{res} ( \mathrm{\textbf{X}}_{i}, \mathrm{\textbf{W}}_{i} ) }{ \partial \mathrm{\textbf{X}}_{l} } \Big) \bigg) \\
&= \frac{ \partial \zeta }{ \partial \mathrm{\textbf{X}}_{L} } \bigg( 1 + \sum_{i=l}^{L-1} \Big( \mathrm{\textbf{U}}_{i} \frac{ \partial {F}_{att} ( \mathrm{\textbf{X}}_{i}, \mathrm{\textbf{U}}_{i} ) }{ \partial \mathrm{\textbf{X}}_{l} } \\
& \qquad \qquad \qquad + {F}_{att} ( \mathrm{\textbf{X}}_{i}, \mathrm{\textbf{U}}_{i} ) \frac{ \partial \mathrm{\textbf{U}}_{i} }{ \partial \mathrm{\textbf{X}}_{l} } \Big) \bigg) \\
&= \frac{ \partial \zeta }{ \partial \mathrm{\textbf{X}}_{L} } \bigg( 1 + \sum_{i=l}^{L-1} \Big( \mathrm{\textbf{U}}_{i} \Big( \frac{ \partial {F}_{att} ( \mathrm{\textbf{X}}_{i}, \mathrm{\textbf{U}}_{i} ) }{ \partial \mathrm{\textbf{U}}_{i} } \frac{ \partial \mathrm{\textbf{U}}_{i} }{ \partial \mathrm{\textbf{X}}_{l} } \\
& \qquad \qquad \qquad \qquad + \frac{ \partial {F}_{att} ( \mathrm{\textbf{X}}_{i}, \mathrm{\textbf{U}}_{i} ) }{ \partial \mathrm{\textbf{X}}_{i} } \frac{ \partial \mathrm{\textbf{X}}_{i} }{ \partial \mathrm{\textbf{X}}_{l} } \Big) \\
& \qquad \qquad \qquad + {F}_{att} ( \mathrm{\textbf{X}}_{i}, \mathrm{\textbf{U}}_{i} ) \frac{ \partial \mathrm{\textbf{U}}_{i} }{ \partial \mathrm{\textbf{X}}_{l} } \Big) \bigg),
\end{aligned}
\end{equation}
so, after sorting out Eq.~\ref{eq48}, we obtain:
\begin{equation}\label{eq49}
\frac{ \partial \zeta }{ \partial \mathrm{\textbf{X}}_{l} } = \frac{ \partial \zeta }{ \partial \mathrm{\textbf{X}}_{L} } \bigg( 1 + \sum_{i=l}^{L-1} \Big( \underbrace{ \beta \cdot \frac{ \partial \mathrm{\textbf{U}}_{i} }{ \partial \mathrm{\textbf{X}}_{l} }}_{ \propto \nabla \mathrm{\textbf{U}}_{i} } + \underbrace{ \gamma \cdot \frac{ \partial \mathrm{\textbf{X}}_{i} }{ \partial \mathrm{\textbf{X}}_{l} }}_{\propto \nabla \mathrm{\textbf{X}}_{i}} \Big) \bigg),
\end{equation}
where the gradients of residual and identity mappings are denoted by $\nabla \mathrm{\textbf{U}}_{i}$ and $\nabla \mathrm{\textbf{X}}_{i}$. By using the strategy of joint modeling for residual and identity flows, the trade-off process of identity and residual mappings can be integrated into the BP optimization, so that residual flows can provide more efficient complementary modeling for identity mappings.

\section{Experiments}
\label{sec5}
In this section, lots of experiments are conducted to verify the performance of the proposed methods. Firstly, the datasets and implementation details of networks are introduced. Secondly, the effects of different types of Inner-Imaging are analyzed, while the ablation study for various subassemblies of the InI module is conducted. Thirdly, comparisons of our models with state-of-the-art results are provided. Finally, we give the discussions of experimental results.

\subsection{Datasets}
\label{sec5.1}
As CIFAR-10, CIFAR-100, SVHN, and ImageNet are often used as benchmark datasets in image recognition experiments. They are also used here.

\textbf{CIFAR-10 and CIFAR-100}~\cite{krizhevsky2009learning}. The two datasets consist of $32 \times 32$ colored images. Both of them contain 60000 images divided equally into 10 and 100 classes. There are 50000 training images and 10000 for testing. The standard data augmentation (translation/mirroring) widely adopted as~\cite{he2016deep} is used for training sets.

\textbf{SVHN}~\cite{netzer2012reading}. The street view house numbers dataset contains $32 \times 32$ colored images of 73257 samples in the training set and 26032 for testing, with 531131 digits for additional training. Here all training data are used without data augmentation.

\textbf{ImageNet}~\cite{deng2009imagenet:}. It is used in ILSVRC 2012, which contains 1.2 million training images, 50000 validation images, and 100000 for testing, with 1000 classes. Standard data augmentation is adopted for the training set, and the $224 \times 224$ crop is randomly sampled. All images are normalized into $[0, 1]$, with mean values and standard deviations.

\subsection{Implementation Details}
\label{sec5.2}
\begin{table}[!t]
\centering
\footnotesize
\caption{Various types of G-filter sets.}
\begin{tabular}{|c|c|p{140pt}<{\centering}|}
\hline
Type & Name & Set of G-filters \\
\hline
\multirow{5}{*}{Square}
& square-$1$ & $\{ (3 \times 3) \}$ \\
& square-$2$ & $\{ (1 \times 1), (3 \times 3) \}$ \\
& square-$3$ & $\{ (1 \times 1), (3 \times 3), (5 \times 5) \}$ \\
& square-$4$ & $\{ (1 \times 1), (2 \times 2), (3 \times 3), (5 \times 5) \}$ \\
& square-$5$ & $\{ (1 \times 1), (2 \times 2), (3 \times 3), (4 \times 4), (5 \times 5) \}$ \\
\hline
\multirow{5}{*}{Mix}
& mix-$1$ & $\{ (3 \times 3) \}$ \\
& mix-$2$ & $\{ (1 \times 5), (3 \times 3) \}$ \\
& mix-$3$ & $\{ (1 \times 5), (3 \times 3), (5 \times 1) \}$ \\
& mix-$4$ & $\{ (1 \times 1), (1 \times 5), (3 \times 3), (5 \times 1) \}$ \\
& mix-$5$ & $\{ (1 \times 1), (1 \times 5), (3 \times 3), (5 \times 1), (5 \times 5) \}$ \\
\hline
Horizontal & horizon-$n$ & $\{ (1 \times 1), \dots, (1 \times n) \}$ \\
\hline
Vertical & vertical-$n$ & $\{ (1 \times 1), \dots, (n \times 1) \}$ \\
\hline
\hline
\multirow{2}{*}{Simplified${\rm ^a}$\tnote{a}}
& simple-$1$ & $\{ (2 \times 1) \}$ \\
& simple-$3$ & $\{ (1 \times 1), (2 \times 1), (2 \times 2) \}$ \\
\hline
Dilated${\rm ^b}$\tnote{b} & d & $\{ \dots, (5 \times 5, s=2) \}$ \\
\hline
\end{tabular}
\begin{tablenotes}
\footnotesize
\item[a]${\rm ^a}$Used only for simplified version of InI module in ResNets.
\item[b]${\rm ^b}$Used only in conjunction with other types of G-filters, not separately.
\end{tablenotes}
\label{table1}
\end{table}

\begin{figure*}[!t]
\centering
\includegraphics[scale=0.58]{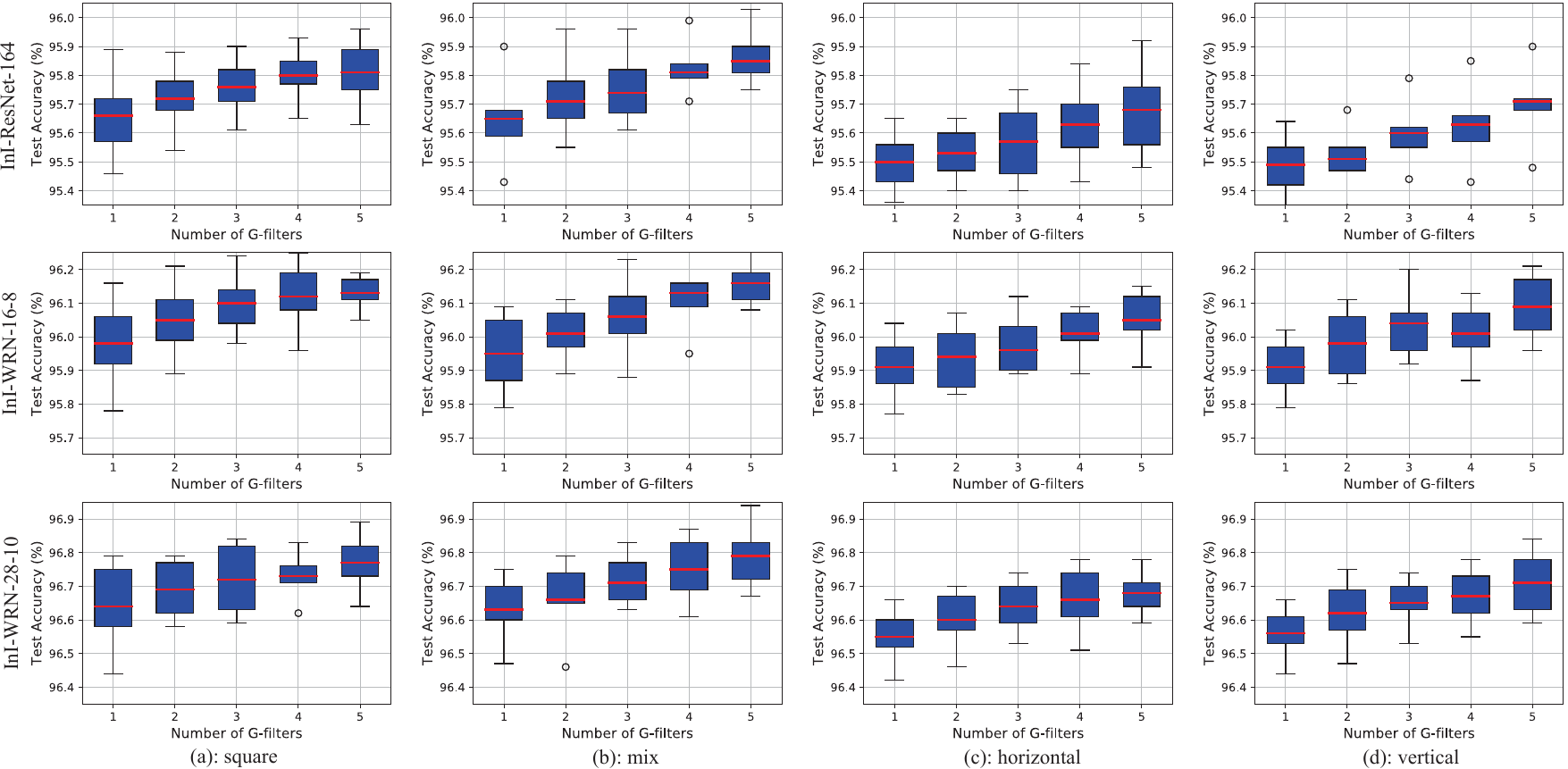}
\caption{Test accuracy curves by InI-ResNet and InI-WRN concerning G-filters with various types on CIFAR-10, the results are reported over $5$ runs. For each column. (a): The square type; (b): The mix type; (c): The horizontal type; (d): The vertical type.}
\label{fig6}
\end{figure*}

\textbf{Networks.} The proposed InI module is applied to several popular CNN networks that are taken as backbones, including All Convolutional Net (All-CNN)~\cite{springenberg2014striving}, Pre-act Residual Network (ResNet)~\cite{he2016identity}, Wide Residual Network (WRN)~\cite{zagoruyko2016wide}, Pyramidal Residual Networks (PyramidNet)~\cite{han2017deep}. All the default settings of the backbones are followed. The InI module is adopted in every block of the backbone networks. For example, ResNet, usually each block contains two convolutional layers. We add the InI module to the last layer of each block. Detailed block settings are described in the references~\cite{springenberg2014striving,he2016identity,zagoruyko2016wide,han2017deep}. The proposed method is also compared with typical channel-wise attention~\cite{hu2018squeeze-and-excitation} using the same backbones. The batch normalization~\cite{ioffe2015batch} is performed following G-filters. Our implementation is based on MXNet and GluonCV\footnote{https://gluon-cv.mxnet.io}

\textbf{Training.} The SGD with 0.9 Nesterov momentum is used to train models on CIFAR, where epoch number is 300 epochs for ResNet and 200 epochs for other backbones, and 80 epochs on SVHN, where the batch-size is 64. For ResNet and All-CNN, the learning rate starts at 0.1 and is divided by 10 at $50\%$, $75\%$ of the number of total epochs, for WRN, it is divided by 5 at 60, 120, 160 epochs. On ImageNet, we train models for 100 epochs with the batch size of 256, the initial learning rate is $0.1$ and reduced by 10 every 30 epochs.

\textbf{Settings of the InI Module.} Since the size of the inner-imaged map needs to be taken into account in the setting of the G-filter, it follows the rules as: if the receptive field of any G-filter exceeds the size of the inner-imaged map in any layer, the G-filter will be automatically discarded.

For the inner-imaged maps, their shapes are defined close to a square: $(n=20, m=C/10)$ for WRN, $(n=8, m=C/8)$ for All-CNN and $(n=2C/16, m=16)$ for other backbones. The effects of size changes of the inner-imaged maps are tested on classification performance. For all kinds of ResNet, the special version of the InI module is used by default, which is introduced in Section~\ref{sec3.2}. It jointly models the identity and residual mappings.

We set the naming rules for the proposed model: 1. the prefix "InI" is used to the name of the model using the Inner-Imaging module; 2. for different G-filter types, the model name is attached with the type name as the suffix, where the type names of G-filters are listed in Table~\ref{table1}.

\subsection{Analysis of Different Inner-Imaging Types}
\label{sec5.3}
\begin{table*}[!t]
\centering
\footnotesize
\caption{Test error ((mean $\pm$ std) \%) of All-CNN, ResNet, SE-ResNet and multiple modes of InI-models over $5$ runs on CIFAR-10 and CIFAR-100. Results that surpass all competing methods are \textbf{bold} and the overall best results are \textcolor[rgb]{1,0,0}{\textbf{red}}.}
\begin{tabular}{|c|p{36pt}<{\centering}|p{36pt}<{\centering}|p{36pt}<{\centering}|p{36pt}<{\centering}|c|c|c|}
\hline
Model & Joint & Aggregation & Fold & Dilated & Params. & CIFAR-10 & CIFAR-100 \\
\hline
All-CNN~\cite{springenberg2014striving} & $-$ & $-$ & $-$ & $-$ & 1.30M & 7.25 & 33.71 \\
\hline
SE-All-CNN~\cite{hu2018squeeze-and-excitation} & $-$ & $-$ & $-$ & $-$ & 1.35M & 6.55 $\pm$ 0.14 & 32.83 $\pm$ 0.21 \\
\hline
InI-All-CNN-square-1 (ours) & $-$ & $-$ & $\surd$ & $-$ & 1.35M & 6.15 $\pm$ 0.13 & 32.15 $\pm$ 0.17 \\
InI-All-CNN-square-3 (ours) & $-$ & $\surd$ & $\surd$ & $-$ & 1.36M & \textbf{6.06 $\pm$ 0.12} & \textbf{32.02 $\pm$ 0.16} \\
\hline
\hline
ResNet-110~\cite{he2016identity} & $-$ & $-$ & $-$ & $-$ & 1.70M & 6.37 & -- \\
ResNet-164~\cite{he2016identity} & $-$ & $-$ & $-$ & $-$ & 1.70M & 5.46 & 24.33 \\
\hline
SE-ResNet-110~\cite{hu2018squeeze-and-excitation} & $-$ & $-$ & $-$ & $-$ & 1.75M & 5.65 $\pm$ 0.15 & 25.79 $\pm$ 0.15 \\
SE-ResNet-164~\cite{hu2018squeeze-and-excitation} & $-$ & $-$ & $-$ & $-$ & 1.95M & 4.79 $\pm$ 0.16 & 22.47 $\pm$ 0.20 \\
\hline
InI-ResNet-110-square-1${\rm ^*}$\tnote{*} (ours) & $-$ & $-$ & $\surd$ & $-$ & 1.70M & 5.46 $\pm$ 0.10 & 25.21 $\pm$ 0.17 \\
InI-ResNet-110-simple-1 (ours) & $\surd$ & $-$ & $-$ & $-$ & 1.75M & 5.35 $\pm$ 0.17 & 25.12 $\pm$ 0.08 \\
InI-ResNet-110-simple-3 (ours) & $\surd$ & $\surd$ & $-$ & $-$ & 1.77M & 5.23 $\pm$ 0.13 & 24.96 $\pm$ 0.21 \\
InI-ResNet-110-square-1 (ours) & $\surd$ & $-$ & $\surd$ & $-$ & 1.76M & 5.20 $\pm$ 0.12 & 24.93 $\pm$ 0.15 \\
InI-ResNet-110-square-3 (ours) & $\surd$ & $\surd$ & $\surd$ & $-$ & 1.77M & 5.16 $\pm$ 0.14 & 24.87 $\pm$ 0.11 \\
InI-ResNet-110-square-3-d (ours) & $\surd$ & $\surd$ & $\surd$ & $\surd$ & 1.77M & \textbf{5.11 $\pm$ 0.10} & \textbf{24.83 $\pm$ 0.09} \\
\hline
InI-ResNet-164-square-1${\rm ^*}$\tnote{*} (ours) & $-$ & $-$ & $\surd$ & $-$ & 1.88M & 4.59 $\pm$ 0.09 & 22.14 $\pm$ 0.17 \\
InI-ResNet-164-simple-1 (ours) & $\surd$ & $-$ & $-$ & $-$ & 1.95M & 4.53 $\pm$ 0.12 & 21.78 $\pm$ 0.19 \\
InI-ResNet-164-simple-3 (ours) & $\surd$ & $\surd$ & $-$ & $-$ & 2.02M & 4.30 $\pm$ 0.14 & 21.66 $\pm$ 0.11 \\
InI-ResNet-164-square-1 (ours) & $\surd$ & $-$ & $\surd$ & $-$ & 1.99M & 4.34 $\pm$ 0.14 & 21.59 $\pm$ 0.16 \\
InI-ResNet-164-square-3 (ours) & $\surd$ & $\surd$ & $\surd$ & $-$ & 2.02M & 4.24 $\pm$ 0.10 & 21.48 $\pm$ 0.18 \\
InI-ResNet-164-square-3-d (ours) & $\surd$ & $\surd$ & $\surd$ & $\surd$ & 2.02M & 4.15 $\pm$ 0.08 & 21.44 $\pm$ 0.15 \\
InI-ResNet-164-mix-5 (ours) & $\surd$ & $\surd$ & $\surd$ & $-$ & 2.04M & 4.15 $\pm$ 0.09 & 21.33 $\pm$ 0.17 \\
InI-ResNet-164-mix-5-d (ours) & $\surd$ & $\surd$ & $\surd$ & $\surd$ & 2.04M & \textcolor[rgb]{1,0,0}{\textbf{4.11 $\pm$ 0.13}} & \textcolor[rgb]{1,0,0}{\textbf{21.31 $\pm$ 0.13}} \\
\hline
\end{tabular}
\begin{tablenotes}
\footnotesize
\item[*]${\rm *}$: without joint modeling of residual and identity mappings in ResNets (same in hereinafter).
\end{tablenotes}
\label{table3}
\end{table*}
\begin{table*}[!t]
\centering
\footnotesize
\caption{Test error ((mean $\pm$ std) \%) of WRN, SE-WRN and multiple modes of InI-models over $5$ runs on CIFAR-10 and CIFAR-100. Results that surpass all competing methods are \textbf{bold} and the overall best results are \textcolor[rgb]{1,0,0}{\textbf{red}}.}
\begin{tabular}{|c|p{36pt}<{\centering}|p{36pt}<{\centering}|p{36pt}<{\centering}|p{36pt}<{\centering}|c|c|c|}
\hline
Model & Joint & Aggregation & Fold & Dilated & Params. & CIFAR-10 & CIFAR-100 \\
\hline
WRN-22-10~\cite{zagoruyko2016wide}& $-$ & $-$ & $-$ & $-$ & 26.80M & 4.44 & 20.75 \\
WRN-28-10~\cite{zagoruyko2016wide}& $-$ & $-$ & $-$ & $-$ & 36.50M & 4.17 & 20.50 \\
\hline
SE-WRN-16-8~\cite{hu2018squeeze-and-excitation}& $-$ & $-$ & $-$ & $-$ & 11.10M & 4.58 $\pm$ 0.12 &  20.94 $\pm$ 0.17 \\
SE-WRN-22-10~\cite{hu2018squeeze-and-excitation}& $-$ & $-$ & $-$ & $-$ & 27.05M & 4.08 $\pm$ 0.13 & 19.55 $\pm$ 0.12 \\
SE-WRN-28-10~\cite{hu2018squeeze-and-excitation}& $-$ & $-$ & $-$ & $-$ & 36.80M & 3.78 $\pm$ 0.19 & 19.03 $\pm$ 0.14 \\
\hline
InI-WRN-16-8-simple-3 (ours) & $\surd$ & $\surd$ & $-$ & $-$ & 11.14M & 4.01 $\pm$ 0.15 & 19.30 $\pm$ 0.11 \\
InI-WRN-16-8-square-1 (ours) & $\surd$ & $-$ & $\surd$ & $-$ & 11.10M & 4.02 $\pm$ 0.13 & 19.29 $\pm$ 0.18 \\
InI-WRN-16-8-square-3 (ours) & $\surd$ & $\surd$ & $\surd$ & $-$ & 11.14M & 3.90 $\pm$ 0.09 & 19.20 $\pm$ 0.13 \\
InI-WRN-16-8-square-3-d (ours) & $\surd$ & $\surd$ & $\surd$ & $\surd$ & 11.14M & 3.85 $\pm$ 0.13 & 19.16 $\pm$ 0.16 \\
InI-WRN-16-8-mix-5 (ours) & $\surd$ & $\surd$ & $\surd$ & $-$ & 11.20M & 3.84 $\pm$ 0.07 & 19.08 $\pm$ 0.19 \\
InI-WRN-16-8-mix-5-d (ours) & $\surd$ & $\surd$ & $\surd$ & $\surd$ & 11.20M & \textbf{3.80 $\pm$ 0.11} & \textbf{19.05 $\pm$ 0.15} \\
\hline
InI-WRN-22-10-simple-3 (ours) & $\surd$ & $\surd$ & $-$ & $-$ & 27.12M & 3.63 $\pm$ 0.12 & 18.38 $\pm$ 0.13 \\
InI-WRN-22-10-square-1 (ours) & $\surd$ & $-$ & $\surd$ & $-$ & 27.10M & 3.62 $\pm$ 0.16 & 18.43 $\pm$ 0.08 \\
InI-WRN-22-10-square-3 (ours) & $\surd$ & $\surd$ & $\surd$ & $-$ & 27.12M & 3.50 $\pm$ 0.18 & 18.36 $\pm$ 0.06 \\
InI-WRN-22-10-square-3-d (ours) & $\surd$ & $\surd$ & $\surd$ & $\surd$ & 27.12M & 3.48 $\pm$ 0.14 & 18.27 $\pm$ 0.09 \\
InI-WRN-22-10-mix-5 (ours) & $\surd$ & $\surd$ & $\surd$ & $-$ & 27.16M & 3.51 $\pm$ 0.10 & 18.25 $\pm$ 0.13 \\
InI-WRN-22-10-mix-5-d (ours) & $\surd$ & $\surd$ & $\surd$ & $\surd$ & 27.16M & \textbf{3.46 $\pm$ 0.14} & \textbf{18.24 $\pm$ 0.09} \\
\hline
InI-WRN-28-10-square-1${\rm ^*}$\tnote{*} (ours) & $-$ & $-$ & $\surd$ & $-$ & 36.60M & 3.47 $\pm$ 0.11 & 18.53 $\pm$ 0.14 \\
InI-WRN-28-10-simple-1 (ours) & $\surd$ & $-$ & $-$ & $-$ & 36.80M & 3.40 $\pm$ 0.13 & 18.30 $\pm$ 0.06 \\
InI-WRN-28-10-simple-3 (ours) & $\surd$ & $\surd$ & $-$ & $-$ & 36.85M & 3.33 $\pm$ 0.06 & 18.28 $\pm$ 0.04 \\
InI-WRN-28-10-square-1 (ours) & $\surd$ & $-$ & $\surd$ & $-$ & 36.82M & 3.36 $\pm$ 0.13 & 18.26 $\pm$ 0.10 \\
InI-WRN-28-10-square-3 (ours) & $\surd$ & $\surd$ & $\surd$ & $-$ & 36.88M & 3.28 $\pm$ 0.10 & 18.08 $\pm$ 0.12 \\
InI-WRN-28-10-square-3-d (ours) & $\surd$ & $\surd$ & $\surd$ & $\surd$ & 36.88M & 3.24 $\pm$ 0.04 & 18.06 $\pm$ 0.15 \\
InI-WRN-28-10-mix-5 (ours) & $\surd$ & $\surd$ & $\surd$ & $-$ & 36.90M & 3.21 $\pm$ 0.09 & 18.02 $\pm$ 0.12 \\
InI-WRN-28-10-mix-5-d (ours) & $\surd$ & $\surd$ & $\surd$ & $\surd$ & 36.90M & \textcolor[rgb]{1,0,0}{\textbf{3.19 $\pm$ 0.10}} & \textcolor[rgb]{1,0,0}{\textbf{17.98 $\pm$ 0.07}} \\
\hline
\end{tabular}
\label{table4}
\end{table*}

\begin{table*}[!t]
\centering
\footnotesize
\caption{Test error((mean $\pm$ std) \%) of InI-models over $5$ runs, compare with state-of-the-art results, using pre-act ResNet, WRN and PyramidNet as backbone, on CIFAR and SVHN. The top-three results are \textcolor[rgb]{1,0,0}{\textbf{red}}, \textcolor[rgb]{0,1,0}{\textbf{green}} and \textcolor[rgb]{0,0,1}{\textbf{blue}} respectively. The FLOPs and Epoch Time is recorded by running on CIFAR-100.}
\begin{tabular}{|c|p{30pt}<{\centering}|p{30pt}<{\centering}|c|p{50pt}<{\centering}|p{40pt}<{\centering}|p{40pt}<{\centering}|p{35pt}<{\centering}|}
\hline
\centering{Model} & Depth & Params. & FLOPs & Epoch Time & CIFAR-10 & CIFAR-100 & SVHN \\
\hline
original ResNet~\cite{he2016deep} & 110 & 1.7M & 253.1M & 37 sec & 6.43 & 25.16 & -- \\
\multirow{2}{*}{pre-act ResNet~\cite{he2016identity}} & 164 & 1.7M & 380.6M & 65 sec & 5.46 & 24.33 & -- \\
 & 1001 & 10.2M & 2.537G & -- & 4.62 & 22.71 & -- \\
\multirow{2}{*}{ResNet Stochastic depth~\cite{huang2016deep}} & 110 & 1.7M & -- & -- & 5.23 & 24.58 & 1.75 \\
 & 1202 & 10.2M & 2.840G & -- & 4.91 & -- & -- \\
\hline
Wide ResNet~\cite{zagoruyko2016wide} & 28 & 36.5M & 5.243G & 112 sec & 4.17 & 20.50 & -- \\
\hline
FractalNet~\cite{larsson2017fractalnet:} & 21 & 38.6M & -- & -- & 5.22 & 23.30 & 2.01 \\
W/dropout \& droppath & 21 & 38.6M & -- & -- & 4.60 & 23.73 & 1.87 \\
\hline
DenseNet~\cite{huang2017densely} & 100 & 27.2M & 13.780G & 96 sec & 3.74 & 19.25 & 1.59 \\
DenseNet-BC ($k=40$) & 190 & 25.6M & 9.388G & -- & 3.46 & 17.18 & -- \\
\hline
ResNeXt~\cite{xie2017aggregated} & 29 & 34.4M & 10.704G & -- & 3.65 & 17.77 & -- \\
\hline
PyramidNet~\cite{han2017deep} & 272 & 26.0M & 8.176G & 174 sec & 3.31 & 16.35 & -- \\
\hline
\multirow{2}{*}{CliqueNet ($k=80 / k=150$)~\cite{yang2018convolutional}} & 15 & 8M & 6.880G & -- & 5.17 & 22.78 & 1.53 \\
 & 30 & 10M & 8.490G & -- & 5.06 & 21.83 & 1.64 \\
\hline
DMRNet-Wide~\cite{zhao2018deep} & 32 & 14.9M & -- & -- & 3.94 & 19.25 & \textcolor[rgb]{0,0,1}{\textbf{1.51}} \\
DMRNet-Wide~\cite{zhao2018deep} & 50 & 24.8M & -- & -- & 3.57 & 19.00 & 1.55 \\
DMRNeXt~\cite{zhao2018deep} & 29 & 26.7M & -- & -- & \textcolor[rgb]{1,0,0}{\textbf{3.06}} & 17.55 & -- \\
\hline
ConDenseNet~\cite{huang2018condensenet} & 160 & 3.1M & 1.084G & -- & 3.46 & 17.55 & -- \\
\hline
InI-ResNet-simple-3 (ours) & 164 & 2.02M & 381.8M & 57 sec & 4.30 & 21.66 & -- \\
InI-ResNet-square-3 (ours) & 164 & 2.02M & 381.9M & 59 sec & 4.24 & 21.48 & -- \\
InI-WRN-simple-3 (ours) & 28 & 36.85M & 5.259G & 126 sec & 3.33 & 18.28 & -- \\
InI-WRN-square-3 (ours) & 28 & 36.88M & 5.262G & 128 sec & 3.28 & 18.08 & \textcolor[rgb]{0,1,0}{\textbf{1.49}} \\
InI-WRN-mix-5 (ours) & 28 & 36.90M & 5.266G & 131 sec & 3.21 & 18.02 & 1.53 \\
InI-WRN-mix-5-d (ours) & 28 & 36.90M & 5.268G & 132 sec & 3.19 & 17.98 & \textcolor[rgb]{1,0,0}{\textbf{1.47}} \\
InI-PyramidNet-square-3 (ours) & 272 & 27.5M & 8.184G & 188 sec & 3.13 & \textcolor[rgb]{0,0,1}{\textbf{15.99}} & -- \\
InI-PyramidNet-mix-5 (ours) & 272 & 27.8M & 8.188G & 191 sec & \textcolor[rgb]{0,0,1}{\textbf{3.11}} & \textcolor[rgb]{0,1,0}{\textbf{15.81}} & -- \\
InI-PyramidNet-mix-5-d (ours) & 272 & 27.8M & 8.121G & 192 sec & \textcolor[rgb]{0,1,0}{\textbf{3.07}} & \textcolor[rgb]{1,0,0}{\textbf{15.77}} & -- \\
\hline
\end{tabular}
\label{table5}
\end{table*}

\textbf{Effects of the shape of G-filter.} As the core component of the InI framework, these issues need to be investigated: whether different G-filter combinations will have a serious impact on model performance (including the number and shape of G-filter), and what strategies can be applied to find the better combinations of G-filters.

As listed in Table~\ref{table1}, a variety of G-filter combinations are designed, including square, slender (vertical and horizontal), and mixed cases. We also introduce the dilated G-filter, and only one type of dilated G-filter is used: $5 \times 5$ receptive field, dilated rate = 2. The dilated G-filter is used only together with other types of G-filters, not individually. These combinations are applied to the InI models so as to observe their performance changes in different settings, as shown in Fig.~\ref{fig6}.

It can be seen from Fig.~\ref{fig6} that the performance of the InI module increases with the number of G-filters. Secondly, the performance of the types of square and mix G-filters is generally better than that of horizontal and vertical G-filters. Particularly, when the number of square G-filters is relatively large, the performance growth rate tends to be flat, and when the number of mixed G-filters is large, better results can be obtained. These indicate that the diversity of G-filter types can bring benefits to the InI model. Moreover, as deduced in Section~\ref{sec4.1}, when we add the $(1 \times 1)$ G-filter to the G-filter set, a more obvious performance improvement can be obtained, which verifies the effect of the special group which only contains one channel, as Eq.~\ref{eq49}.

Fig.~\ref{fig5} can help us analyze the above results, which illustrates the corresponding channel distribution of groups obtained by using G-filters with various shapes, on the original channel sequence. In the group of channels formed by horizontal and vertical G-filters, the former only contains adjacent channels, while the latter contains channels with slightly distant intervals. These monotonous grouping strategies lead to their mediocre performance. In the channel group formed by square G-filter, there are adjacent channels and distant channels. However, as the size of G-filter increases, the coverage of square G-filter becomes wider and wider, too many square G-filters also reduce the diversity of channel groups. In contrast, the mixed set G-filter composed of horizontal, vertical, and square G-filters is more effective.

It can also be observed in Fig.~\ref{fig5}(d) that the dilated~\cite{yu2016multi-scale} G-filter can use fewer parameters than square G-filter to complete a large scope of channel scanning while overcoming the redundancy caused by a high overlap rate between channel groups. Besides, the analysis of the benefits of the dilated G-filter is given in the appendices.

Overall, combining G-filters with multiple sizes and shapes can indeed bring better modeling capabilities to the InI mechanisms. The effect on the performance of different G-filter type selection is minimal. Although the maximum performance of InI can be achieved by combining G-filters of various shapes as much as possible, we can get competitive performance with G-filter type "square-3-d". In appendices, we also analysis the effects of the shape of inner-imaged map.

\subsection{Ablation Studies}
\label{sec5.4}
\begin{figure}[!ht]
\centering
\includegraphics[scale=0.56]{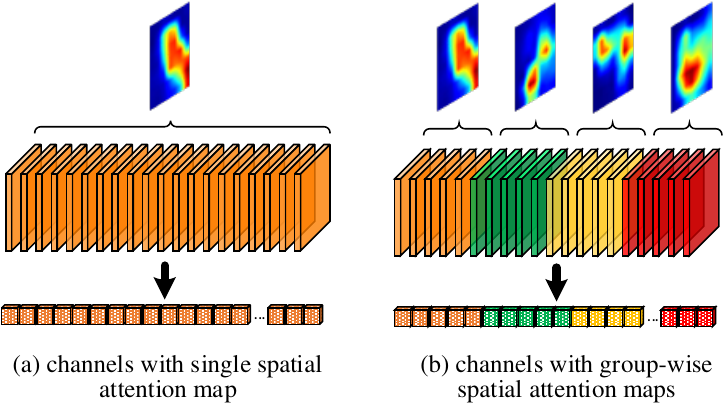}
\caption{The illustrations of spatial attention. (a): Single spatial attention; (b): Group-wise multiple spatial attention.}
\label{fig8}
\end{figure}
\begin{figure*}[!t]
\centering
\includegraphics[scale=0.56]{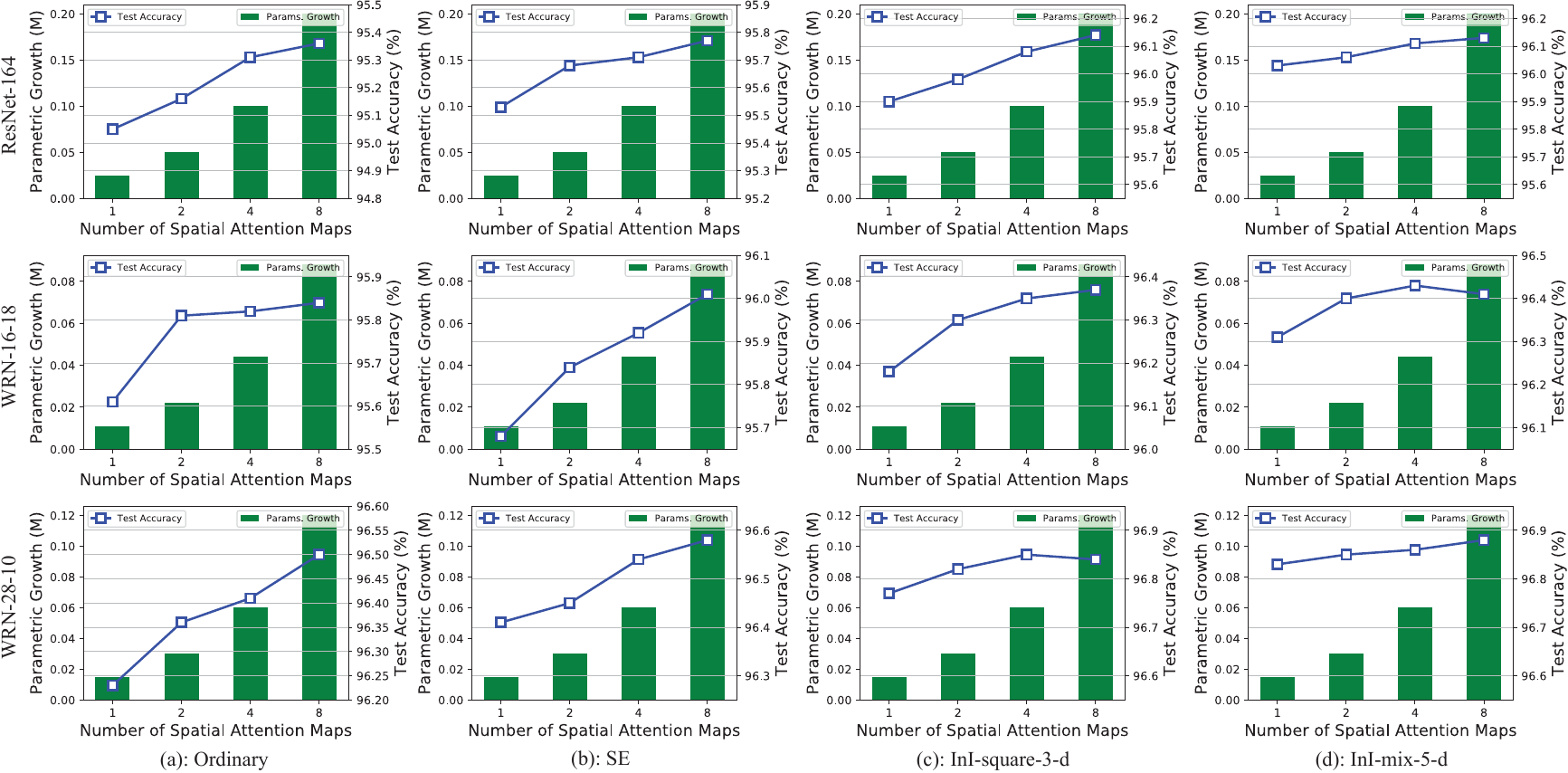}
\caption{The amount of parameter growth and test accuracy concerning the number of spatial attention maps with various networks under modes of (a): ordinary; (b) SE; (c) InI-square-3 and (d) InI-mix-5-d on CIFAR-10.}
\label{fig9}
\end{figure*}

We study the effects of all tricks used in the InI models, which include: (1) Jointly modeling from identity and residual mappings (Joint); (2) Multi-shape G-filter aggregation (Aggregation); (3) Folded inner-imaged map (Fold); (4) Add dilated G-filter. These tricks are gradually added to the same backbone.

As listed in Tables~\ref{table3} and~\ref{table4}, the best records with the same backbone are in bold, and the best results are highlighted in red. The variable 'Params' in all tables refers to the parameter amount of each model. As the components are gradually added, the test error continues to decrease, and the fully configured InI-models achieved the best results. Besides, many small-scale InI-models achieve performance close to or even better than the larger-scale typical models. Every trick in our InI mechanism shows the improvements in classification results.

It can be found that the well-configured (with three or more proposed tricks) InI models can improve the typical channel-wise attention network by nearly 0.6\%. Compared with SE-Net, InI-models can bring more performance improvement.

\subsection{Coordination with Spatial Attention}
\label{sec5.5}
The spatial attention mechanism~\cite{woo2018cbam:} is a strategy to adjust the pixel-level weights on the feature maps dynamically. In this section, the cooperative ability of the proposed InI mechanism for spatial attention models is validated.

We use the InI module after conducting the spatial attentional operation as follows, and the spatial attention map (SP map) is recorded as $\xi$,
\begin{equation}\label{eq50}
\mathrm{\textbf{U}}_{l}^{new} = \mathrm{\textbf{s}} \circ {F}_{spa} \left( \mathrm{\textbf{U}}_{l}, \xi \right) = \left[ \xi \circ \mathrm{\textbf{u}}_{l}^{1}, \dots, \xi \circ \mathrm{\textbf{u}}_{l}^{C} \right]
\end{equation}
where ${F}_{spa}(\cdot)$ is the function of spatial attention and $\mathrm{\textbf{U}}_{l}^{new}$ is the final feature matrix in layer $l$.

As shown in Fig.~\ref{fig8}, multiple SP maps $\{ {\xi}^{1}, \dots, {\xi}^{\tau} \}$ are also introduced on several divided channel groups to highlight the function of InI model, as:
\begin{equation}\label{eq51}
\begin{aligned}
\mathrm{\textbf{U}}_{l}^{new} &= \mathrm{\textbf{s}} \circ {F}_{spa} \left( \mathrm{\textbf{U}}_{l}, \{ {\xi}^{1}, \dots, {\xi}^{\tau} \} \right) \\
&= \left[ {\xi}^{1} \circ \mathrm{\textbf{u}}_{l}^{1}, \dots, {\xi}^{1} \circ \mathrm{\textbf{u}}_{l}^{ \frac{C}{\tau} }, \dots \right. \\
& \qquad \qquad \left. , {\xi}^{\tau} \circ \mathrm{\textbf{u}}_{l}^{C - \frac{C}{\tau} + 1}, \dots, {\xi}^{\tau} \circ \mathrm{\textbf{u}}_{l}^{C} \right]
\end{aligned}
\end{equation}

Fig.~\ref{fig9} shows the growth of parameters and test accuracy on CIFAR-10 when using different numbers of SP maps. It is noted that the InI model fits well with the spatial attention mechanism and gets better results than other methods. As the number of SP maps increases, the increments of parameters rise exponentially, and the performance of the model also improves. Still, when the number of SP maps reaches 8, the performance growth becomes less noticeable or even slightly declined.

The possible reason is that too many SP maps lead to a bit over-fitting on CIFAR-10 composed with small pictures.

In the appendices\footnote{also available at https://github.com/scut-aitcm/Appendices-of-InI-Net}, we provide further comparison results of the proposed InI models with more attention-based models.

\subsection{Comparison Results}
\label{sec5.6}
\begin{table}[!ht]
\centering
\footnotesize
\caption{Single crop error rates ($\%$) on ImageNet.}
\begin{tabular}{|c|c|c|c|}
\hline
Model & Params. & top-1 & top-5 \\
\hline
ResNet-50~\cite{he2016deep} & 25.60M & 24.70 & 7.80 \\
ResNet-101~\cite{he2016deep} & 44.60M & 23.60 & 7.10 \\
ResNet-152~\cite{he2016deep} & 60.30M & 23.00 & 6.7 \\
DenseNet-121~\cite{huang2017densely} & 7.98M & 25.02 & 7.71 \\
CliqueNet~\cite{yang2018convolutional} & 14.38M & 24.01 & 7.15 \\
\hline
SE-ResNet-50~\cite{hu2018squeeze-and-excitation} & 28.10M & 23.29 & 6.62 \\
SE-ResNet-101~\cite{hu2018squeeze-and-excitation} & 49.40M & 22.38 & 6.07 \\
SE-ResNet-152~\cite{hu2018squeeze-and-excitation} & 65.50M & 21.57 & 5.73 \\
\hline
InI-ResNet-50-mix-5-d (ours) & 28.77M & 22.10 & 5.79 \\
InI-ResNet-101-mix-5-d (ours) & 50.58M & \textbf{21.33} & \textbf{5.28} \\
\hline
\hline
CBAM-ResNet-50~\cite{woo2018cbam:} & 28.16M & 22.66 & 6.31 \\
CBAM-ResNet-101~\cite{woo2018cbam:} & 49.48M & 21.51 & 5.69 \\
\hline
InI-ResNet-50-mix-5-d + spa (ours) & 28.83M & 21.44 & 5.57 \\
InI-ResNet-50-mix-5-d + spa$\times$4 (ours) & 29.00M & 21.16 & 5.42 \\
InI-ResNet-101-mix-5-d + spa (ours) & 50.66M & 20.81 & 5.17 \\
InI-ResNet-101-mix-5-d + spa$\times$4 (ours) & 50.89M & \textbf{20.52} & \textbf{5.06} \\
\hline
\end{tabular}
\label{table6}
\end{table}

Table~\ref{table5} reports the comparison results between the InI-models and state-of-the-art networks, where 'FLOPs' denotes the float-point computation amount. 'Epoch Time' indicates the average training time per epoch (devices info is given in the appendices). We only list the Epoch Time records of comparison models we have repeat implemented. It can be noticed that the InI-model obviously improves the performance of the baseline method, and the InI-PyramidNet-mix-5-d outperforms state-of-the-art results and achieves the best results on CIFAR-100 and achieves the second-best performance on CIFAR-10. Although the results of the InI-PyramidNet-square-3 are slightly lower, they achieve the third-best and second-best results on CIFAR-10 and CIFAR-100. The InI-WRN-square-3, InI-WRN-mix-5, and InI-WRN-mix-5-d are also very competitive, and two of them meet the best and second-best performance on SVHN. Also, we note that DMRNeXt enhances the interaction and complementarity of the residual flow and the constant flow on the two channels. This scheme of forcibly grouping convolution channels and improving communication between groups has achieved the best performance on the CIFAR-10 dataset after combining with ResNeXt. However, on the more challenging datasets, the proposed InI-Net all performed the best records.

The proposed InI-models have a general and relatively significant performance improvement for the corresponding backbone networks, requiring only little additional parameters and computations. The InI-model has a better convolutional component organization; it makes convolution kernels own the diversity, complementarity, and overall completeness, thus avoiding the redundancy of feature maps and enhancing the integrity of feature representation.

The experimental results on ImageNet are listed in Table~\ref{table6}, and the best results are highlighted in bold. It can be seen that InI-ResNet-50-mix-5-d improve the top-1 and top-5 error rate than SE-ResNet-50 by $1.21\%$ and $0.85\%$, respectively. InI-ResNet-101-mix-5-d exceeds SE-ResNet-101 by $1.04\%$ and $0.8\%$ respectively, which is even better than the performance of SE-ResNet-152. Compared the networks with mechanisms of both channel-wise attention and spatial attention, the InI-ResNet-mix-5-d without spatial attention strategy can obtain better performance than CBAM-ResNet with spatial attention. The results of our InI-ResNet-mix-5-d + spa $\times$ 4 improve that of CBAM-ResNet by $0.9\% \sim 1.5\%$ on top-1 and $0.6\% \sim 0.9\%$ on top-5. Similarly, the smaller InI-models still defeat the larger compared models. The details about the computation and time cost of the experimental models are put in the appendices.

Fig.~\ref{fig10} illustrates some thermal visualization results of feature maps on ImageNet with different models. It can be found that the InI model can focus more accurately and entirely on recognition objects.

The above results on ImageNet show the InI model also works well in large-scale image recognition tasks. In the appendices, we also show the experimental results of the InI models on the object detection task.

\begin{figure}[!ht]
\centering
\includegraphics[scale=0.33]{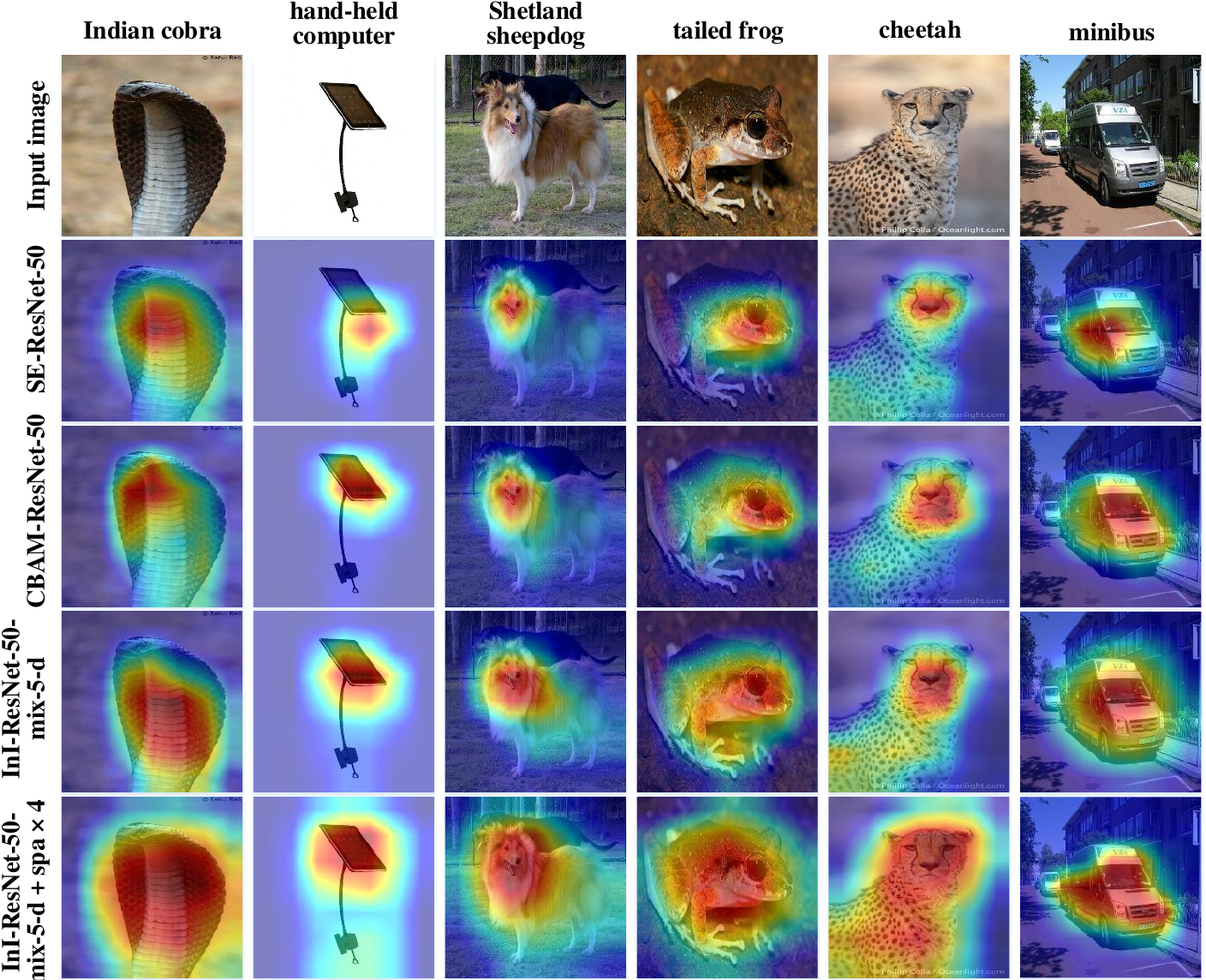}
\caption{The Grad-CAM~\cite{selvaraju2017grad-cam:} visualization for different models with the backbone of ResNet-50, on ImageNet.}
\label{fig10}
\end{figure}

\subsection{Discussions}
\label{sec5.7}
Our experimental results show that the InI model can improve the modeling ability of CNN, especially for smaller convolution structures. The InI mechanism can adequately stimulate their potential and make the classification results close to or even exceed the original larger CNN models. Besides, the InI model only needs very few additional parameters.

The performance improvement of the InI model than the traditional channel-wise attention validates the necessity of channel group relational modeling and the superior modeling ability of the InI model for diversified channel relationships.

In experiments, many possibilities for the implementation of the InI mechanism are discussed. However, it does not mean that the InI model needs to rely on many hyper-parametric adjustments. On the contrary, the InI model is insensitive to hyper-parameters, such as the G-filter type. Conventional G-filter selection and the simple combination can obtain very competitive performance, illustrating that the InI model is very robust. Also, the InI model is highly scalable and flexible. The various patterns and combinations of the InI model can be extended to pursue extreme excelsior modeling performance.

The InI model has high universality so that it can be applied to any CNN structure. It also has good adaptability to other enhancement mechanisms for CNNs, such as spatial attention.

\section{Conclusion}
\label{sec6}
In this paper, we propose the Inner-Imaging architecture for convolutional networks, which present a novel strategy to model the channel relationships in CNNs. The proposed Inner-Imaging architecture uses the convolutional G-filter to organize the grouping relations of the channels, explicitly models the channel in-group coordination and the complementary inter-group ties. This design effectively improves the modeling efficiency of convolutional networks. The proposed method is easy to use and extensible, and its superior performance is verified on multiple benchmark datasets.

In future work, we will test the effectiveness of the proposed InI mechanism on more CNN architectures and apply more data augmentation strategies, like Auto Augmentation~\cite{cubuk2018autoaugment} and Mixup~\cite{zhang2018mixup}, to further improve the image recognition performance of the proposed models. Also, in recent years, neural architecture search (NAS)~\cite{elsken2019neural} has become a new trend in the design of neural network structures. We will integrate the "Inner-imaging" mechanism into the paradigm of NAS. When searching for the novel CNN architectures, we plan to give CNN models independent detection ability of grouped channel relations and optimize them.


%



\section*{Acknowledgment}
\label{sec7}
\footnotesize{This study was supported by Guangdong Province Key Area R \& D Plan Project (2020B1111120001, 2018B010107002), China National Science Foundation (Grant Nos. 61273363, 61976092, 61722205, 61751205, 61751202, U1611461), Guangzhou Science and Technology Planning Project (Grant No. 201604020179 and 201803010088), and Natural Science Foundation of Guangdong (Grant No. 2018A030313356).}

\ifCLASSOPTIONcaptionsoff
  \newpage
\fi



%
\bibliography{references}{}
\bibliographystyle{IEEEtran}

\normalsize

\noindent \textbf{Yang Hu}
\textbf{(S'19)} received the MA.Eng degree from Kunming University of Science and Technology in 2016. He is currently pursuing the Ph.D degree in South China University of Technology, China. His research interests include neural network and deep learning, biomedical information processing.

\

\noindent \textbf{Guihua Wen}
received the Ph.D. degree in Computer Science and Engineering, South China University of Technology, and now is professor, doctoral supervisor at the School of Computer Science and Technology of South China University of Technology. His research area includes Cognitive affective computing, Machine Learning and data mining. He is also professor in chief of the data mining and machine learning laboratory at the South China University of Technology.

\

\noindent \textbf{Mingnan Luo}
Mingnan Luo is currently a master candidate in the College of Computer Science and Engineering, South China University of Technology. His main research interests include image processing and deep learning.

\

\noindent \textbf{Dan Dai}
received the MA.Eng degree from Kunming University of Science and Technology in 2016. She is currently pursuing the Ph.D degree in South China University of Technology, China. Her research interests include machine learning and biomedical information processing.

\

\noindent \textbf{Wenming Cao}
received the M.S. degree from the School of Automation, Huazhong University of Science and Technology, Wuhan, China, in 2015. He is currently pursuing the Ph.D. degree with the Department of Computer Science, City University of Hong Kong, Hong Kong. His current research interests include data mining and machine learning.

\

\noindent \textbf{Zhiwen Yu}
\textbf{(S'06-M'08-SM'14)} is a senior member of IEEE, ACM, IRSS, CCF (China Computer Federation) and CAAI (Chinese Association for Artificial Intelligence). Dr. Yu obtained the PhD degree from City University of Hong Kong in 2008. The research areas of Dr. Yu focus on data mining, machine learning, bioinformatics and pattern recognition. Until now, Dr. Yu has been published more than 100 referred journal papers and international conference papers, including TKDE, TEC, TCYB, TMM, TCVST, TCBB, TNB, INS, PR, Bioinformatics, SIGKDD, and so on. Please refer to the homepage for more details: http://www.hgml.cn/yuzhiwen.

\

\noindent \textbf{Wendy Hall}
DBE, FRS, FREng is Regius Professor of Computer Science at the University of Southampton, and is the executive director of the Web Science Institute. She became a Dame Commander of the British Empire in the 2009 UK New Year's Honours list, and is a Fellow of the Royal Society. She has previously been President of the ACM, Senior Vice President of the Royal Academy of Engineering, and a member of the UK Prime Minister's Council for Science and Technology. She is also co-chair of the UK government's AI Review.





\end{document}